\documentclass{article}

\usepackage{comment}
\usepackage{microtype}
\usepackage{graphicx}
\usepackage{subcaption}
\usepackage{booktabs} 
\usepackage{multirow}
\usepackage{xcolor} 
\usepackage{makecell}

\usepackage{hyperref}

\usepackage[accepted]{icml2024}

\usepackage{amsmath}
\usepackage{amssymb}
\usepackage{mathtools}
\usepackage{amsthm}
\usepackage{amsmath}
\usepackage{svg}
\usepackage{multirow}
\usepackage[normalem]{ulem}
\useunder{\uline}{\ul}{}
\usepackage{array}
\usepackage{float}

\usepackage[
    backend=biber,
    style=nature,
]{biblatex}
\let\cite=\supercite

\usepackage[capitalize,noabbrev]{cleveref}

\theoremstyle{plain}

\theoremstyle{definition}

\theoremstyle{remark}

\newcommand\startsupplement{%
    \makeatletter 
       \setcounter{table}{0}
       \renewcommand{\thetable}{S\arabic{table}}
       \setcounter{figure}{0}
       \renewcommand{\thefigure}{S\arabic{figure}}
    \makeatother}

\icmltitlerunning{ExChanGeAI: An End-to-End Platform and Efficient Foundation Model for Electrocardiogram Analysis and Fine-tuning}

\begin{document}

\twocolumn[
    \icmltitle{ExChanGeAI: An End-to-End Platform and Efficient Foundation Model for Electrocardiogram Analysis and Fine-tuning}
    
    \begin{icmlauthorlist}
    \icmlauthor{Lucas Bickmann}{yyy}
    \icmlauthor{Lucas Plagwitz}{yyy}
    \icmlauthor{Antonius Büscher}{yyy,yyz}
    \icmlauthor{Lars Eckardt}{yyz}
    \icmlauthor{Julian Varghese}{yyy}
    \end{icmlauthorlist}

    \icmlaffiliation{yyy}{Institute of Medical Informatics, University Münster, Germany}
    \icmlaffiliation{yyz}{Clinic for Cardiology II: Electrophysiology, University Hospital Münster, Germany}
    
    \icmlcorrespondingauthor{Lucas Bickmann}{lucas.bickmann@uni-muenster.de}
        
    \icmlkeywords{Health Informatics, Electrocardiogram, Machine Learning, Deep Learning, Foundation Model, End-to-End Platform}
    
    \vskip 0.3in
]



\printAffiliationsAndNotice{}  

\begin{abstract}
Electrocardiogram data, one of the most widely available biosignal data, has become increasingly valuable with the emergence of deep learning methods, providing novel insights into cardiovascular diseases and broader health conditions. However, heterogeneity of electrocardiogram formats, limited access to deep learning model weights and intricate algorithmic steps for effective fine-tuning for own disease target labels result in complex workflows. In this work, we introduce \textit{ExChanGeAI}, a web-based end-to-end platform that streamlines the reading of different formats, pre-processing, visualization and custom machine learning with local and privacy-preserving fine-tuning. ExChanGeAI is adaptable for use on both personal computers and scalable to high performance server environments. The platform offers state-of-the-art deep learning models for training from scratch, alongside our novel open-source electrocardiogram foundation model \textit{CardX}, pre-trained on over one million electrocardiograms. Evaluation across three external validation sets, including an entirely new testset extracted from routine care, demonstrate the fine-tuning capabilities of ExChanGeAI. CardX outperformed the benchmark foundation model while requiring significantly fewer parameters and lower computational resources. The platform enables users to empirically determine the most suitable model for their specific tasks based on systematic validations.
The code is available at \href{imigitlab.uni-muenster.de/published/exchangeai}{this URL}.

{\footnotesize
\qquad \textit{Keywords - }{\small\textbf{Health Informatics, Electrocardiogram, Machine Learning, Deep Learning, Foundation Model, End-to-End Platform}}
}
\end{abstract}
\section{Introduction}
Deep Learning methods applied for Electrocardiography (ECG) analyses have demonstrated their potential as practice-changing diagnostic tools, providing critical insights to heart related diseases\cite{lin_ai-enabled_2024,adedinsewo_artificial_2024,sau_artificial_2024}. While the push for newer technologies and improved performance metrics is essential, ensuring these advancements are accessible for general use is equally important. Tools like ChatGPT have demonstrated the potential for broad and easy application of AI, allowing users to leverage sophisticated technologies without in-depth expertise. Clinician-researchers who seek to apply machine learning to assess its potential benefits should have access to solutions that facilitate exploration and application without requiring extensive technical knowledge from data handling to data analysis. To address this, an intuitive platform could streamline  ECG data-loading, preprocessing, analysis and fine-tuning of models for specific tasks while effectively leveraging prior knowledge embedded in large pre-trained models. Moreover, existing pre-trained models for ECG data may not disclose model weights, which presents significant challenges (see related work below). While open weights empower users to utilize and adapt the model, they also promote reproducibility and transparency\cite{widder_why_2024,polevikov_advancing_2023}. 
Additionally, the broad use of medical data is crucial for the advancement of personalised and specialised medicine but inhibits some immediate risks such as data breaches\cite{rahman_impact_2024}. As datasets continue to grow and come from diverse sources, ensuring data security becomes increasingly complex. To address this challenge, specialized decentralized learning techniques such as federated learning or swarm learning allow valuable insights to be gained without directly sharing sensitive data\cite{Rauniyar_federated}. Furthermore, establishing uniform data standards can simplify data handling, reduce technical barriers related to varying data formats, and eliminate the need for programming, thereby making advanced analytics accessible to a broader range of users.

In this work, we introduce a novel open-source end-to-end platform for 12-lead ECGs called ExChanGeAI that integrates the following processes: (1) data loading and preprocessing of multiple input formats, (2) manual and computer aided analysis of ECG waveform data, (3) one-click fine-tuning of classification models, allowing users to train and customise ML models with no prior expertise, (4) the trained models utilize the cross-platform industry-standard Open Neural Network Exchange (ONNX), enabling deployment in every instance of ExChanGeAI and facilitating the exchange of custom models across different instances, and (5) prediction of diseases using the pre-trained and fine-tuned models. Model sharing is supported via an integrated and adaptable WebDav file-server called Model ExChanGe. Among the provided pretrained models, we provide the novel foundation ECG model Cardiology eXpert (CardX). The platform is built upon the principle of open-source and -weights, offering full transparency and control, empowering users to understand, fine-tune, and contribute to the advancement of ECG analysis models.
\subsection{Related Work}
Multiple studies and reviews have addressed electrocardiogram classification, and have shown that fine-tuning and transfer learning improve classification results\cite{jang_effectiveness_2021,weimann_transfer_2021,chato_survey_2023}. A study has reported improved model accuracy by fine-tuning networks trained on diverse datasets, demonstrating enhanced performance transitioning to smaller datasets\cite{avetisyan_deep_2024}. However, used data and pretrained models were not shared. Another study used transfer learning with convolutional neural networks (CNNs) for atrial fibrillation classification, pretraining on large public datasets and fine-tuning on smaller sets, achieving performance gains\cite{weimann_transfer_2021}. While code was available, pretrained models were not shared, and usability remains a significant barrier. Multiple reviews have summarized ECG analysis pipelines and deep learning methods; such as detailed essential pipeline steps\cite{kaplan_berkaya_survey_2018}, and reviews of techniques like CNNs and RNNs for arrhythmia classification\cite{ebrahimi_review_2020}. The SelfONN model\cite{qin_lightweight_2024} showed competitive performance in general ECG classification on PTB-XL but lacked resource sharing. Various types of autoencoders, including low-rank attention\cite{zhang_ecg_2024}, long short-term memory (LSTM)\cite{roy_ecg-net_2023}, adversarial\cite{thambawita_deepfake_2021}, and denoising\cite{singh_attention-based_2022} approaches, have been explored for feature extraction, anomaly detection, and noise handling. The low-rank attention autoencoder reported high accuracy on two datasets by focusing on spatial features. ECG-NET, based on LSTM, proclaimed high accuracy for arrhythmia classification on a single database in beat-based validation. An adversarial autoencoder with a temporal CNN published superior scores of anomaly detection for two datasets. The attention-based denoising autoencoder improved noisy ECG signal reconstruction. However, limitations across these studies include dataset dependence, restricted generalizability, lack of publicly available pretrained models and code, and validation variability.

In a recent study leveraging the gold-standard PTB-XL\cite{goldberger_physiobank_2000,wagner_ptb-xl_2020} dataset, the performance characteristics of multiple deep learning models were evaluated across a spectrum of training-data sizes\cite{bickmann_post_2023}. Findings indicated that the Inception- and XceptionTime architectures\cite{ismail_fawaz_inceptiontime_2020,rahimian_xceptiontime_2020} exhibited particularly compelling performances. Specifically, InceptionTime demonstrated superior efficacy when trained with smaller datasets, whereas XceptionTime surpassed all other models in performance as training dataset size increased. This suggests a potential trade-off between model complexity and data requirements for optimal diagnostic accuracy in this domain. Due to the demonstrated strength in low- and high-data scenarios, these leading architectures for ECG analysis are highly relevant for evaluation and inclusion in the platform, particularly in contexts where training data availability may vary, such as in medical contexts. 

There have been several claimed foundation models in the field of ECG classification\cite{han_foundation_2024}. To the best of our knowledge, these however, are trained on a singular database\cite{li_electrocardiogram_2024} and are yet undisclosed, or have closed source code and weights\cite{wang_anyecg_2025} in general. In one case the published weights are different from the original model of the paper due to privacy concerns\cite{mckeen_ecg-fm_2024}. A request to publish another trained model has been declined due to intellectual property and legal concerns\cite{mathew_foundation_2024}. They are trained with techniques such as contrastive- and masked-learning. This allows for unsupervised training, however restricts the learning to the latent space. For downstream tasks, such as classification, fine-tuning is required. The publications report high scores for classification, however additional tasks are not available in the published model. While these models mark significant progress in the field, they often grapple with issues such as overfitting to specific datasets, limited scalability, or insufficient handling of the variability and quality complications intrinsic to diverse ECG datasets. 

By providing an accessible, containerized platform that requires no pre-existing machine learning expertise, ExChanGeAI facilitates the visualization, transformation, prediction, and fine-tuning designed for ECG data. It can leverage pre-trained models, and it supports a broad range of formats and preprocessing steps, ensuring usability across different clinical and research settings. Within ExChangeAI, we provide our novel and open source foundation model CardX, trained on over one million ECGs from six distinct sources. CardX serves as a valuable resource for researchers, enabling efficient training and fine-tuning of deep learning models while preserving data privacy. This enhances both the accessibility and utility of advanced ECG analysis.
\section{Results}
\begin{figure*}[!htb]
\vspace{1em}
    \centering
    \begin{subfigure}{0.49\linewidth}
        \includegraphics[width=\linewidth]{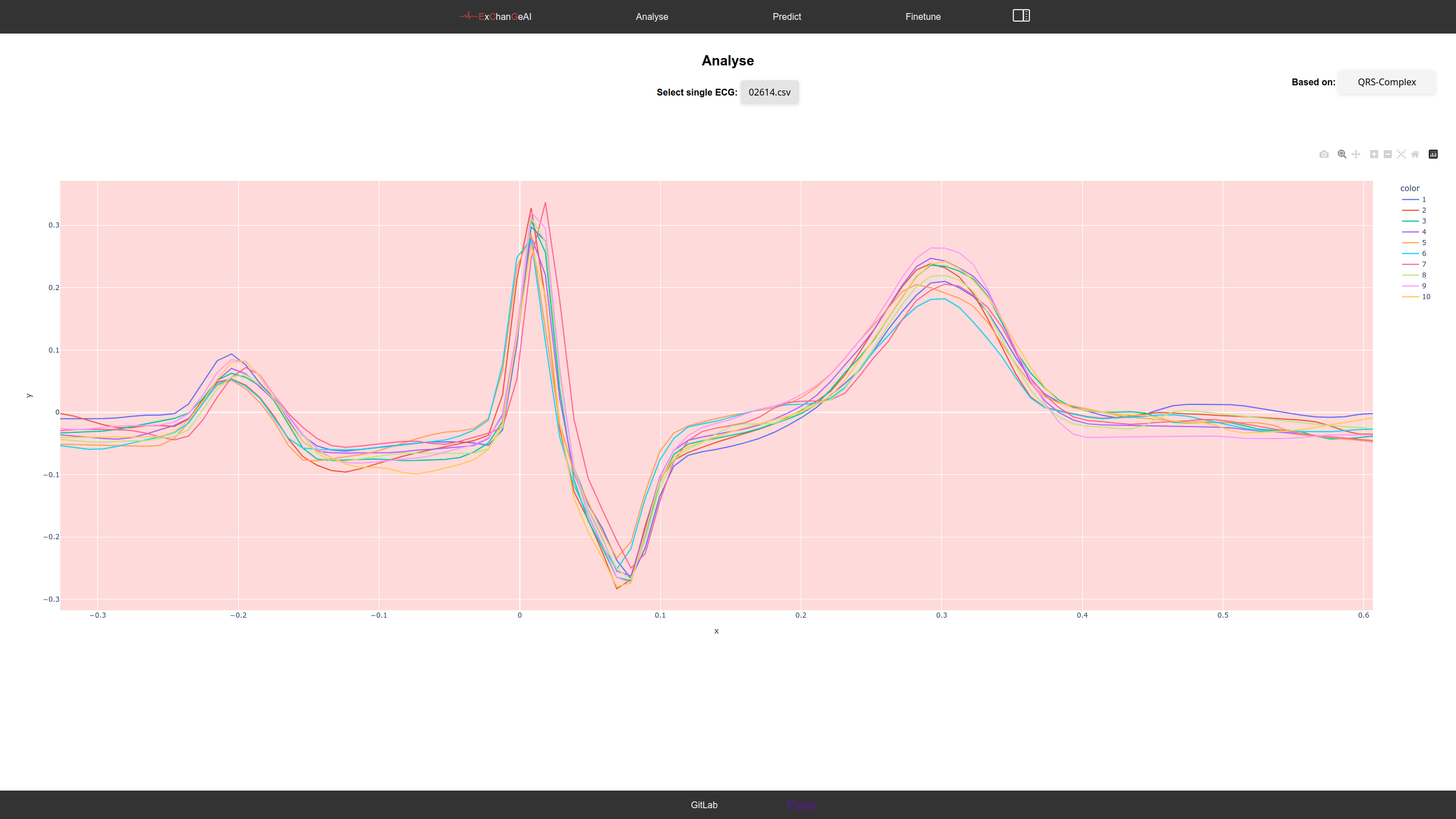}
        \caption{Analyse page.}
    \end{subfigure}
    \hfill
    \begin{subfigure}{0.49\linewidth}
        \includegraphics[width=\linewidth]{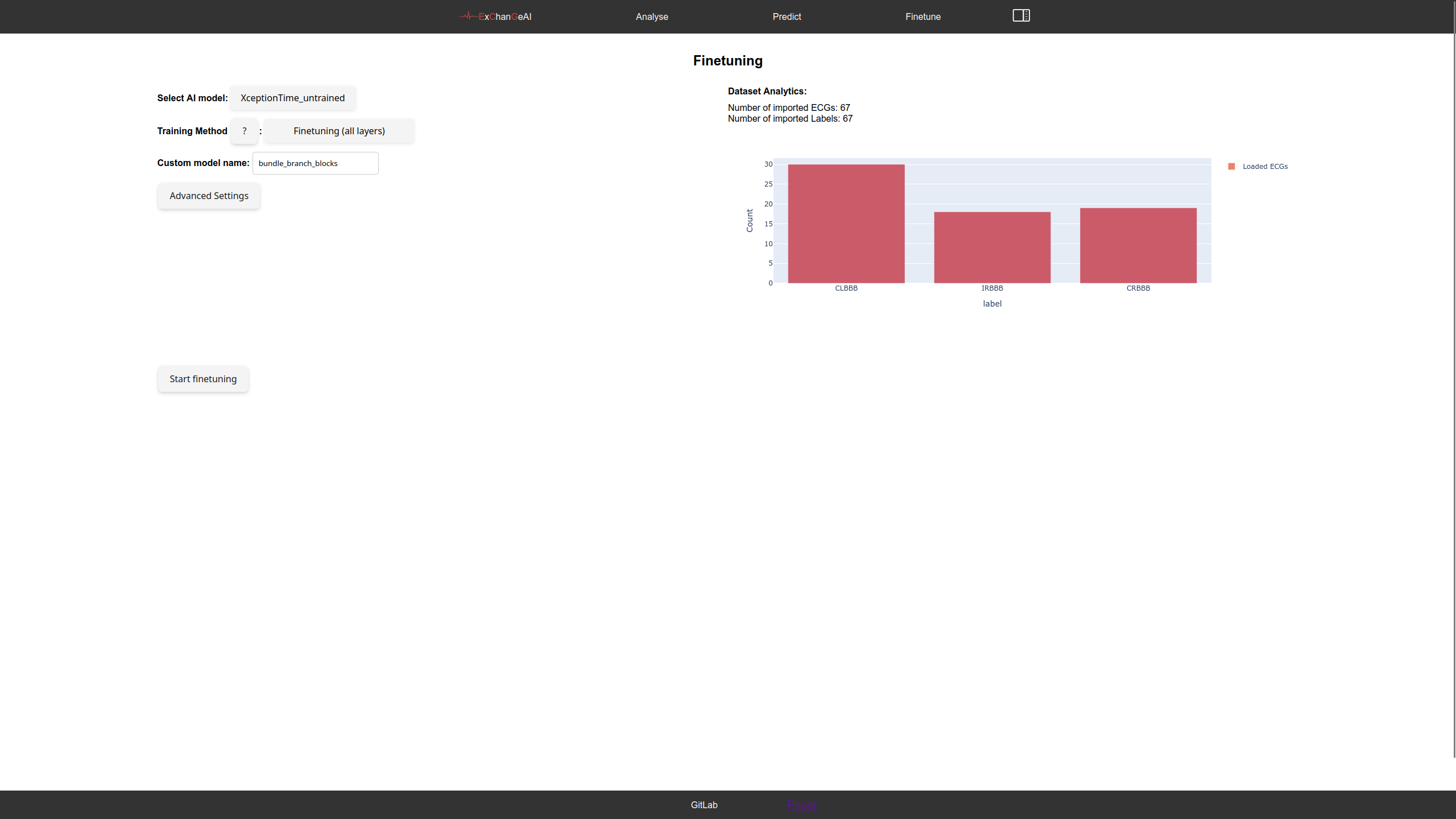}
        \caption{Finetuning page.}
    \end{subfigure}
    \caption{Analyse (left) and Finetuning (right) page of ExChanGeAI with sample data.}
    \label{fig:platform}
\end{figure*}
The interface is visually structured into different foci, such as data analysis and finetuning. Figure \ref{fig:platform} shows the view for analysis and fine-tuning. The analysis module provides interactive visualization of individual ECG files. Users can select to view ECG data based on multiple views - raw time series, QRS complexes, fiducial point annotation, Rlign median beats, and Rlign time-aligned ECG. When visualizing QRS complexes, the interface displays overlaid waveforms, potentially highlighting morphological features. For raw time series visualization, the platform presents the standard 12-lead ECG signals as separate plots, allowing for detailed inspection of each lead's waveform. The fine-tuning view displays options for model selection, training method, and custom model naming. A bar chart visualization summarizes the distribution of labels within the loaded dataset, presenting counts for categories, such as 'CLBBB', 'IRBBB', and 'CRBBB' as shown in the example. Numerical dataset characteristics, including the total number of imported ECGs and labels, are presented as well.

To evaluate ExChanGeAI's capabilities and assess our novel foundation model CardX for systematic downstream performances, we fine-tuned various architectures within ExChanGeAI on classification tasks, using PTB-XL for intra-dataset evaluation and three external test sets - Yang et al.\cite{lai_practical_2023,yang_offline_2023}, MIMIC-IV-ECG\cite{gow_mimic-iv-ecg_nodate}, and the newly generated  Emergency Department Münster (EDMS) - to assess inter-dataset generalizability (Table \ref{tab:results}). In comparison to ExChangeAI models, we have evaluated the Data Science \& Artificial Intelligence Laboratory Seoul National University Physionet 2021 model (DSAIL SNU)\cite{reyna_will_2021,reyna_issues_2022,han_towards_2021} and the foundation model ECG-FM\cite{mckeen_ecg-fm_2024}. The evaluation spans different diagnostic categories (myocardial infarction, ST/T-Changes, conduction disturbance, hypertrophy) and specific comparisons (anterior vs. inferior myocardial infarction, bundle branch block subtypes, and revascularization), using F1 score, robustness metrics (interquartile range (IQR), coefficient of variation (CV)), and computational scaling (parameters, floating point operations per second (FLOPs)). Moreover, the evaluation compares pre-trained and fine-tuned foundation models on the inter-datasets, benchmarking against training models from scratch using the ExChanGeAI platform. To demonstrate the advanced classification task revascularization (“does the patient require revascularization?”), which is not available in PTB-XL, the models are trained and fine-tuned using the new EDMS dataset. In a special case to showcase the capability of cross-task transfer learning, the pretrained XceptionTime model (superclasses with PTB-XL folds 1-8) was also separately fine-tuned.

The foundation model ECG-FM is the largest with over 90 million parameters, followed by this work`s Cardiology eXpert (CardX) (15M), DSAIL SNU (2M), Xception- (401K) and InceptionTime (457K). In terms of computational complexity, ECG-FM is the most demanding (14 GFLOPS), followed in descending order by Inception- (460 MFLOPS), XceptionTime (256 MFLOPS), CardX (201 MFLOPS), and DSAIL SNU (89 MFLOPS). Performance across inter-datasets is presented in table \ref{tab:results}. \cref{supp:weightedf1-internal,supp:macrof1-internal,supp:macrof1-external} in the supplement provide weighted F1 score for intra- and macro F1 scores for both intra- and inter-dataset evaluations. The tables are structured column-wise by model and row-wise by classification task and dataset. Specifically, rows represent the classification tasks across different test sets (Yang et al., MIMIC-IV, EDMS). To illustrate, the first row presents superclass classification scores on the  Yang et al. dataset, where baseline XceptionTime achieves an F1 score of 0.335 and fine-tuned XceptionTime reaches 0.647. As an example for myocardial infarction, the InceptionTime model achieves the second-best F1 score across both datasets (.726 on MIMIC-IV and .584 on EDMS), while XceptionTime achieves a slightly higher score on the former (.734), and CardX on the latter (.625). The last rows show the aggregated statistics, showing Xception- and InceptionTime have the best average and median F1 scores, outlining the top overall performing models, while CardX and DSAIL SNU show the best IQR and CV values, exhibiting the most robust scores across external datasets. 
\begin{table*}[ht]
\vspace{1em}
\small
\renewcommand{\arraystretch}{1.1}
\setlength{\tabcolsep}{4pt}
    \centering
\begin{tabular}{|ll|cccccccc|}
\hline
\multicolumn{2}{|l|}{Model} &
  \multicolumn{2}{c|}{XceptionTime} &
  \multicolumn{1}{c|}{InceptionTime} &
  \multicolumn{1}{c|}{CardX\textdaggerdbl} &
  \multicolumn{2}{c|}{DSAIL SNU} &
  \multicolumn{2}{c|}{ECG-FM} \\ \hline
\multicolumn{2}{|l|}{Training Platform} &
  \multicolumn{5}{c|}{ExChanGeAI} &
  \multicolumn{3}{c|}{custom training code} \\ \hline
\multicolumn{2}{|l|}{Trainable Parameters} &
  \multicolumn{2}{c|}{401K} &
  \multicolumn{1}{c|}{457K} &
  \multicolumn{1}{c|}{15M} &
  \multicolumn{2}{c|}{2M} &
  \multicolumn{2}{c|}{90M} \\ \hline
\multicolumn{2}{|l|}{Finetuned Parameters} &
  \multicolumn{3}{c|}{-} &
  \multicolumn{1}{c|}{1026} &
  \multicolumn{2}{c|}{266} &
  \multicolumn{2}{c|}{1538} \\ \hline
\multicolumn{2}{|l|}{Total FLOPS↓} &
  \multicolumn{2}{c|}{246M} &
  \multicolumn{1}{c|}{460M} &
  \multicolumn{1}{c|}{{\ul 201M}\textdaggerdbl} &
  \multicolumn{2}{c|}{\textbf{89M}} &
  \multicolumn{2}{c|}{14G} \\ \hline
\multicolumn{2}{|l|}{Preprocessing} &
  \multicolumn{3}{c|}{-} &
  \multicolumn{1}{c|}{Moving median} &
  \multicolumn{2}{c|}{Min-max} &
  \multicolumn{2}{c|}{Z-sccore} \\ \hline
\multicolumn{2}{|l|}{Pretrained} &
  \multicolumn{3}{c|}{-} &
  \multicolumn{1}{c|}{\begin{tabular}[c]{@{}c@{}}6 sources\\semi-supervised\end{tabular}} &
  \multicolumn{2}{c|}{\begin{tabular}[c]{@{}c@{}}7 sources\\supervised\end{tabular}} &
  \multicolumn{2}{c|}{\begin{tabular}[c]{@{}c@{}}2 sources, unsupervised\\ Physionet 2021, supervised\end{tabular}} \\ \hline
\multicolumn{2}{|l|}{Task} &
  \multicolumn{3}{c|}{De novo Training} &
  \multicolumn{2}{c|}{Finetune} &
  \multicolumn{2}{c|}{\begin{tabular}[c]{@{}c@{}}\textit{Pretrained}\\\textit{Physionet 2021}\end{tabular}} &
  Finetune \\ \hline
\multicolumn{2}{|l|}{PTB-XL folds} &
  \multicolumn{1}{c|}{1-8} &
  \multicolumn{4}{c|}{9} &
  \multicolumn{2}{c|}{-} &
  9 \\ \hline
\multicolumn{2}{|l|}{Finetuned layers} &
  \multicolumn{3}{c|}{all} &
  \multicolumn{2}{c|}{head} &
  \multicolumn{2}{c|}{-} &
  head \\ \hline
\multicolumn{1}{|l|}{\multirow{3}{*}{Superclasses}} &
  Yang et al. &
  \multicolumn{1}{c|}{.335} &
  \multicolumn{1}{c|}{\textbf{.647}} &
  \multicolumn{1}{c|}{.064} &
  \multicolumn{1}{c|}{.644} &
  \multicolumn{1}{c|}{{\ul .585}} &
  \multicolumn{1}{c|}{\textit{.402}} &
  \multicolumn{1}{c|}{\textit{.491}} &
  .173 \\ \cline{2-10} 
\multicolumn{1}{|l|}{} &
  MIMIC-IV &
  \multicolumn{1}{c|}{.371} &
  \multicolumn{1}{c|}{\textbf{.374}} &
  \multicolumn{1}{c|}{{\ul .358}} &
  \multicolumn{1}{c|}{.208} &
  \multicolumn{1}{c|}{.323} &
  \multicolumn{1}{c|}{\textit{.055}} &
  \multicolumn{1}{c|}{\textit{.192}} &
  .355 \\ \cline{2-10} 
\multicolumn{1}{|l|}{} &
  EDMS &
  \multicolumn{1}{c|}{.432} &
  \multicolumn{1}{c|}{\textbf{.387}} &
  \multicolumn{1}{c|}{{\ul .379}} &
  \multicolumn{1}{c|}{.263} &
  \multicolumn{1}{c|}{.331} &
  \multicolumn{1}{c|}{\textit{.039}} &
  \multicolumn{1}{c|}{\textit{.333}} &
  .216 \\ \hline
\multicolumn{1}{|l|}{\multirow{2}{*}{Myocardial Infarcts}} &
  MIMIC-IV &
  \multicolumn{1}{c|}{.753} &
  \multicolumn{1}{c|}{\textbf{.734}} &
  \multicolumn{1}{c|}{{\ul .726}} &
  \multicolumn{1}{c|}{.502} &
  \multicolumn{1}{c|}{.336} &
  \multicolumn{2}{c|}{\textit{-}} &
  .403 \\ \cline{2-10} 
\multicolumn{1}{|l|}{} &
  EDMS &
  \multicolumn{1}{c|}{.566} &
  \multicolumn{1}{c|}{.484} &
  \multicolumn{1}{c|}{{\ul .584}} &
  \multicolumn{1}{c|}{\textbf{.625}} &
  \multicolumn{1}{c|}{.343} &
  \multicolumn{2}{c|}{\textit{-}} &
  .532 \\ \hline
\multicolumn{1}{|l|}{\multirow{3}{*}{Bundle branch blocks}} &
  Yang et al. &
  \multicolumn{1}{c|}{.730} &
  \multicolumn{1}{c|}{.101} &
  \multicolumn{1}{c|}{\textbf{.792}} &
  \multicolumn{1}{c|}{.610} &
  \multicolumn{1}{c|}{.496} &
  \multicolumn{1}{c|}{\textit{.007}} &
  \multicolumn{1}{c|}{\textit{.089}} &
  {\ul .617} \\ \cline{2-10} 
\multicolumn{1}{|l|}{} &
  MIMIC-IV &
  \multicolumn{1}{c|}{.825} &
  \multicolumn{1}{c|}{\textbf{.820}} &
  \multicolumn{1}{c|}{\textbf{.819}} &
  \multicolumn{1}{c|}{{\ul .571}} &
  \multicolumn{1}{c|}{.333} &
  \multicolumn{1}{c|}{\textit{.000}} &
  \multicolumn{1}{c|}{\textit{.028}} &
  .087 \\ \cline{2-10} 
\multicolumn{1}{|l|}{} &
  \multirow{2}{*}{EDMS} &
  \multicolumn{1}{c|}{.739} &
  \multicolumn{1}{c|}{\textbf{.827}} &
  \multicolumn{1}{c|}{{\ul .732}} &
  \multicolumn{1}{c|}{\textbf{.826}} &
  \multicolumn{1}{c|}{.622} &
  \multicolumn{1}{c|}{\textit{.000}} &
  \multicolumn{1}{c|}{\textit{.118}} &
  .248 \\ \cline{1-1} \cline{3-10} 
\multicolumn{1}{|l|}{Revascularization} &
   &
  \multicolumn{1}{c|}{.750*} &
  \multicolumn{1}{c|}{\textbf{.688}} &
  \multicolumn{1}{c|}{{\ul .645}} &
  \multicolumn{1}{c|}{.614} &
  \multicolumn{1}{c|}{.635} &
  \multicolumn{2}{c|}{\textit{-}} &
  .603 \\ \hline
\multicolumn{1}{|l|}{\multirow{4}{*}{Weighted F1}} &
  Average$\uparrow$ &
  \multicolumn{1}{c|}{.611} &
  \multicolumn{1}{c|}{{\ul .562}} &
  \multicolumn{1}{c|}{\textbf{.567}} &
  \multicolumn{1}{c|}{.540} &
  \multicolumn{1}{c|}{.445} &
  \multicolumn{2}{c|}{\textit{-}} &
  .359 \\ \cline{2-10} 
\multicolumn{1}{|l|}{} &
  Median$\uparrow$ &
  \multicolumn{1}{c|}{.73} &
  \multicolumn{1}{c|}{\textbf{.647}} &
  \multicolumn{1}{c|}{{\ul .645}} &
  \multicolumn{1}{c|}{.610} &
  \multicolumn{1}{c|}{.343} &
  \multicolumn{2}{c|}{\textit{-}} &
  .355 \\ \cline{2-10} 
\multicolumn{1}{|l|}{} &
  IQR$\downarrow$ &
  \multicolumn{1}{c|}{.318} &
  \multicolumn{1}{c|}{.347} &
  \multicolumn{1}{c|}{.353} &
  \multicolumn{1}{c|}{\textbf{.123}} &
  \multicolumn{1}{c|}{{\ul .252}} &
  \multicolumn{2}{c|}{-} &
  .316 \\ \cline{2-10} 
\multicolumn{1}{|l|}{} &
  CV$\downarrow$ &
  \multicolumn{1}{c|}{.308} &
  \multicolumn{1}{c|}{.433} &
  \multicolumn{1}{c|}{.443} &
  \multicolumn{1}{c|}{{\ul .358}} &
  \multicolumn{1}{c|}{\textbf{.310}} &
  \multicolumn{2}{c|}{-} &
  .538 \\ \hline
\end{tabular}
    \caption{Performance evaluation of different models on ECG abnormality classification tasks on the inter-test-datasets. The table compares XceptionTime, InceptionTime, DSAIL SNU, ECG-FM and our foundation model CardX, highlighting the number of parameters, FLOPS , applied preprocessing, pretrained status, weighted F1 score across various fine-tuning tasks, and targets. For each classification target, the best-performing model is highlighted in bold, while the second-best is underlined. Values within a 1\% range are considered equivalent. The pretrained XceptionTime is shown as baseline comparison, but not taken into account for evaluating best performing model, due to the uneven large amount of training data. \textdaggerdbl Internal snapshot of CardX at 200 Epochs, which is a sparsely activated model; counting largest possible activation. *Transfer learning based on pretrained XceptionTime (superclasses with folds 1-8).}
    \label{tab:results}
\end{table*}
XceptionTime and InceptionTime, representing architectures trained de novo on PTB-XL, meaning with random initialization and without any prior training at all, often achieved the highest results across classification tasks. In contrast, the pre-trained models, DSAIL SNU and ECG-FM, exhibited a more nuanced performance profile in our limited data setting. Initially, both models demonstrated suboptimal classification accuracy, especially on datasets outside of their pre-training domain (Physionet 2021). Fine-tuning them on PTB-XL led to significant improvements for both DSAIL SNU and ECG-FM, however, they were outperformed by the de novo InceptionTime (8 out of 9) and XceptionTime models on (7 out of 9) tasks within our evaluation. Our novel foundation model, CardX, developed within the ExChanGeAI platform, outperforms the other foundation ECG-FM, across (7 out of 9) tasks and DSAIL SNU (6 out of 9). Notably, CardX achieves this with significantly reduced computational demands, demonstrating an approximate 70-fold reduction in FLOPs and six-times less parameters compared to ECG-FM. Furthermore, CardX showed higher robustness across diverse datasets, evidenced by the lowest IQR and second lowest CV of weighted F1 scores (.123 and .358). The DSAIL SNU model achieved similar robustness, however with a significant drop in Average and Median F1 scores. While CardX did not consistently achieve the absolute highest F1 scores compared to XceptionTime and InceptionTime, it secured the best average and median score across pre-trained models. 
\section{Discussion}
Our evaluation of ExChanGeAI and the foundation model CardX alongside established architectures, reveals several key insights into model selection, particularly in data-constrained scenarios. The end-to-end platform simplifies both training and fine-tuning, yielding effective performance metrics across diverse ECG classification tasks. Our evaluation shows that training on limited data and testing on inter-datasets inherently implies generalization and performance differences, compared to commonly used intra-dataset evaluations. Consequently, the near-perfect accuracy metrics - in intra-testset and simple tasks, such as tachy- and bradycardia prediction\cite{mckeen_ecg-fm_2024} - are not reproducible when models are evaluated on external, independent datasets. However, when models are evaluated using intra-dataset testing and ample training data is available, achieving high scores becomes more feasible and reproducible on external datasets (see baseline XceptionTime model in Tables \cref{supp:weightedf1-internal,supp:macrof1-internal}). As expected and in line with previous findings\cite{martinez-selles_current_2023,mcdermott_reproducibility_2021}, all models exhibited performance degradation when evaluated on external datasets.

Notably, XceptionTime models exhibit parameter efficiency while achieving competitive performance, underscoring the inherent strengths of this architecture as a previous study has shown\cite{bickmann_post_2023}. However, performance variability was observed across different classification tasks and datasets, indicating a sensitivity to data scaling and parameter optimization within specific model architectures. Pre-trained models, while anticipated to leverage their extensive prior knowledge, presented a mixed picture in our limited data setting. While fine-tuning improved their performance, they generally did not consistently surpass the de novo trained XceptionTime and InceptionTime models. However, the novel foundation model CardX did exhibit enhanced robustness against performance degradation across external datasets in most cases, compared to de novo trained models. It demonstrated the lowest performance variance across all datasets and models, highlighting its robustness across different tasks. Furthermore, CardX consistently achieved higher classification scores than its foundational counterpart ECG-FM, while also offering a significant advantage in computational efficiency, with an approximate 70-fold reduction in FLOPS. As training data availability increases for downstream tasks, we anticipate further performance gains for both model families; however, CardX's superior parameter and computational efficiency indicate a higher potential for scalability in resource-constrained applications. 
In summary, while pre-trained models offer potential, their benefits are not guaranteed in data-constrained scenarios. Untrained architectures can be surprisingly effective. CardX emerges as a promising alternative, outperforming other pre-trained models, offering computational efficiency, and demonstrating superior robustness, highlighting the importance of considering these trade-offs in data-limited clinical environments.
The public availability of the pre-trained and fine-tuned models within the ExChanGeAI platform fosters open science and collaborative model sharing within the research community. ExChanGeAI`s ease of model integration promotes the exchange of readily deployable models, eliminating complexities associated with model-specific installations and runtime dependencies, thereby democratizing access to advanced deep learning techniques. By encapsulating state-of-the-art configurations within a user interface, ExChanGeAI effectively reduces the practical barriers to machine learning implementation, particularly benefiting non-experts and general-purpose applications. While acknowledging potential limitations for expert users seeking highly specialized customizations, the platform`s modular design allows for the future integration of additional compatible architectures, expanding its versatility. Ultimately, ExChanGeAI aims to enhance the accessibility of deep learning model fine-tuning and reduce operational overhead, facilitating broader adoption and accelerating progress in ECG analysis. The inherent influence of model architecture on performance, coupled with the relative consistency of the subsequent training process across architectures, underscores the value of an end-to-end platform that simplifies exploration and deployment of diverse, yet effective, models. Limitations are mainly poised by the available data and infrastructure, even tough  fine-tuning, on recent customary machines, becomes significantly easier due to the large increase in computational power in the current century and wider adoption of specialised hardware, such as graphics and neural processing units.

In conclusion, this work introduced ExChanGeAI, a novel open-source platform designed to streamline and democratize the application of deep learning to ECG analysis. ExChanGeAI simplifies the complex workflow associated with ECG data processing, model training, and fine-tuning. Our evaluation highlights the platform's effectiveness in training and deploying various deep learning architectures, including our novel ECG foundation model CardX, demonstrating a robust balance of performance and computational efficiency, outperforming foundation model competitors. Crucially, our findings underscore that while pre-trained models offer advantages, they are not universally superior in data-constrained scenarios. Indeed, models trained from scratch within ExChanGeAI, such as XceptionTime and InceptionTime, frequently achieved top performance, emphasizing the importance of empirical validation and model selection tailored to specific tasks and datasets.

\section{Methods}
In this section, the two key aspects of this work are presented: (1) the interoperable and standardized platform ExChanGeAI, and (2) the open-source foundation model CardX. The end-to-end platform enables analysis, diagnosis prediction, and model fine-tuning without the need for expert knowledge. The foundation model introduces a novel Mixture of Architecture design, and a new pre-training method utilizing time-aligned electrocardiograms.
\subsection{ExChanGeAI Platform}
\begin{figure*}[!ht]
    \centering
    \includegraphics[width=0.9\linewidth]{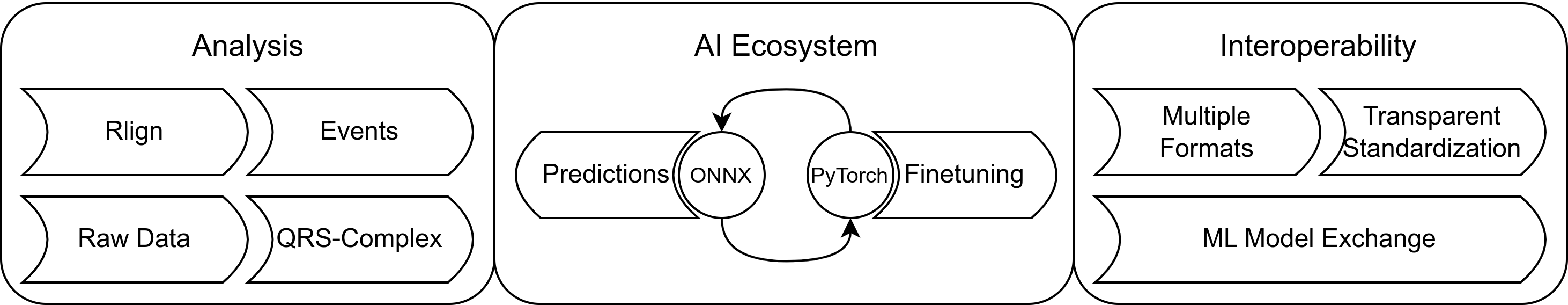}
    \caption{Overview of the end-to-end platform ExChanGeAI and its three main distinct parts: 1. Analysis, 2. The AI Ecosystem and 3. Interoperability. All parts are transparent to the user, and interact without manual interaction.}
    \label{fig:overview}
\end{figure*}
The containerized application ExChanGeAI provides a web interface offering various AI-driven ECG analysis tools for researchers and clinicians. It allows the integration of human analysis and AI predictions within the same end-to-end platform (Figure \ref{fig:overview}). The functionalities include: (1) Analysis, (2) Prediction, (3) Model Exchange, and (4) Training and Fine-tuning. It is open-source, uses the standardized Open Neural Network Exchange (ONNX) model format, and additionally supports and encourages the open-weights practice of machine learning models\cite{widder_why_2024}, with an integrated machine learning model exchange. The platform consists of multiple views with different foci. The analysis view integrates the visualisation of waveforms on raw signals, QRS-complexes, and events (fiducial points), computed transparently by Neurokit2\cite{makowski_neurokit2_2021}. Additionally, precise R-Peak alignment and median beat transformations are supported through the integration of the recently published ECG-preprocessing package, Rlign\cite{plagwitz_rlign_2024}. These data transformations can be exported in different formats for further research. The platform focuses on resting 12-lead ECGs, displayed in a 2x6 grid in mV-scale, and integrates general spatial transformations, including zooming with synchronized adaption across all leads. This signal view has been designed in collaboration with cardiologists for their everyday use. For visualizing QRS-complexes and events, the lead II is conventionally applied. The prediction view uses selected models to predict diagnoses and other targets - such as QTc - based on raw signal data, offering a table with predicted diagnose probabilities (or arbitrary keys), highlighted with a clear colour coding scheme. Models are provided within the platform using an integrated and interchangeable file-server, which enables the crucial aspect of model sharing (see section Model ExChanGe). Currently, we provide multiple models for four targets - Superclasses, anterior and inferior myocardial infarction, diverse bundle branch blocks, and revascularization (see section Results). Researchers can also train a specialist model, based upon their own labeled data. The training and fine-tuning requires no prior knowledge of machine learning. Currently only 12-lead ECG data and corresponding labels are required. ExChanGeAI supports various ECG input formats, including all possible sampling rates, CSV-files (.csv), NumPy-arrays (.npy, .npz), DICOM-stored WaveformSequences (.dcm), Matlab formatted data (.mat), general DAT-files (.dat) and XML-files (.xml). These include the standards of Physionet (DAT), UKBiobank (XML) and the most commonly used formats of other platforms. Internally, all data is normalized by resampling frequency signals to a unified target of 100 Hz using the Fast Fourier Transform. Previous work has shown that the sampling rate does not notably decrease performance of machine learning models, but reduces computational overhead manifold\cite{salimi_exploring_2023}. Signals not conforming to the standardized 12-lead, ten-second waveform are adjusted via expansion or cropping, and all scales are automatically normalized to milliVolt (mV) with a 1000 analog to digital converter units gain (ADC), if necessary (Figure \ref{fig:norm-flowchart}). The platform applies multithreading, asynchronous views and supports the use of CUDA-enabled GPUs for finetuning, including multi GPU support via data parallel training.
\subsection{Model ExChanGe}
The platform enables training of new models in a secure and privacy preserving manner. Still, as seen with many publications in the medical domain, the models are not made public26,28 or depend on external libraries and require specific versions\cite{mckeen_ecg-fm_2024}. To promote the open and interoperable machine learning standard, this work adopts the Open Neural Network Exchange (ONNX) as the primary utilized format. Therefore, our platform is compatible with all ONNX models, honouring the current operation set (Opset $\leq$ 20), and PyTorch models, if the given model structure is provided alongside. The models are not specified with any special requirements, except for a dynamic batch size export. With the use of ONNX, this work aims to ensure that the trained model is widely accessible and interoperable. Therefore, a model exchange is integrated into the platform, where pretrained models are automatically synced and made available for prediction as well as fine-tuning. Additional models can be published into the curated repository, or an own WebDav-instance can be set up. 
By default, ExChanGeAI provides three baseline model architectures - each as pre-trained and untrained models. This includes the XceptionTime\cite{rahimian_xceptiontime_2020}, InceptionTime\cite{ismail_fawaz_inceptiontime_2020}, and the PhysioNet/CinC Challenge 2021 12-lead second best model DSAIL SNU\cite{goldberger_physiobank_2000,reyna_will_2021,han_towards_2021}, due to unavailable weights of the leading model. The latter is initially adapted for ONNX export and initialized with the unavailable co-input features (age, sex) based on the provided default values in its publication. Additional models can be incorporated by uploading into the platform, or the use of the default model exchange file-server. Additionally, we incorporate a novel foundation ECG model (CardX), to extend the research community with a novel open-source and -weights of a large scale pretrained model (see section CardX below). 
\subsection{Fine-tuning Platform}
\begin{figure}[ht]
\vspace{1em}
    \centering
    \includegraphics[width=\linewidth]{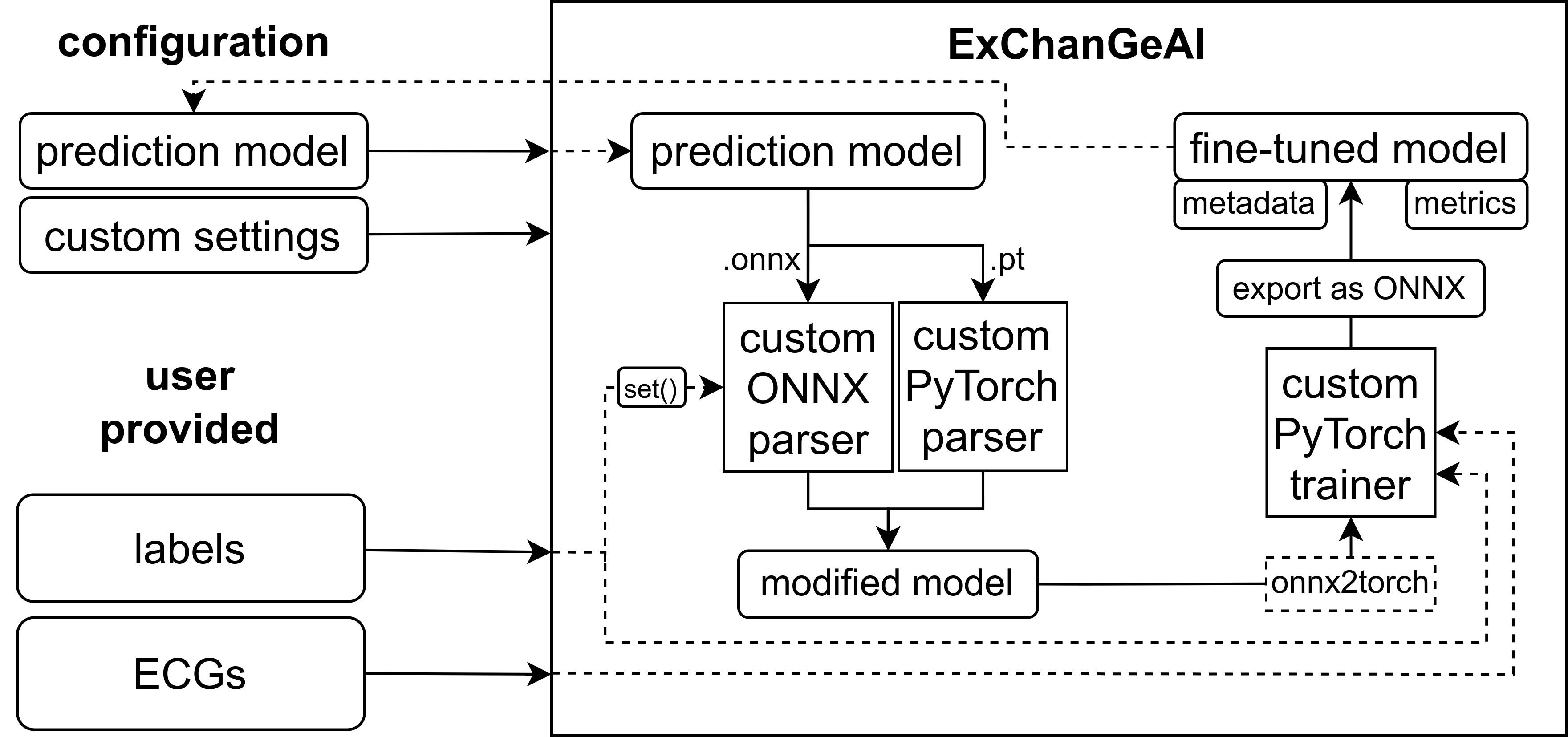}
    \caption{Overview of the semi-automatic fine-tuning process. The user provides labels and 12-lead ECGs, and can change the default configuration if necessary. The backend process is executed without requiring any previous programming knowledge.}
    \label{fig:overview-fine-tuning}
\end{figure}

ExChanGeAI provides a user interface for fine-tuning with user-supplied data and facilitates the exchange of prediction models among researchers. To ensure broad compatibility with various models and platforms, we natively support PyTorch and developed a custom parser for ONNX models to adapt the computation graph where applicable. ExChanGeAI is built upon PyTorch\cite{ansel_pytorch_2024}, which facilitates on-device training, and leverages the ONNX framework during inference. ONNX’s custom training runtime does not support all PyTorch operators with corresponding gradient implementations, such as functions like ReduceMin and diverse Pooling Operators (Opset $\leq18$). While defining custom operators may solve this, it requires detailed implementation knowledge for each unknown operator, creating a significant barrier for practical applications. To overcome these limitations, we employ onnx2torch\cite{developers_onnx2torch_2021} to convert our uniquely parsed ONNX models into PyTorch models. This conversion enhances compatibility, allowing training across different ONNX Opset versions and utilizing the extensive featureset of PyTorch. These conversion steps are computed independently for each fine-tuning process and are entirely transparent to the user (Figure \ref{fig:overview-fine-tuning}). We provide two distinct fine-tuning methods: fine-tuning the classification heads only or training the entire model. Both methods utilise pre-trained weights, if a pre-trained model is selected. The fine-tuning process employs the AdamW optimizer\cite{loshchilov_decoupled_2018}, and the ExponentialLR learning rate scheduler ($\gamma=0.9$). Initially, a learning rate finder is executed to automatically adjust the initial learning rate based on the provided model and data. It has been shown that this method improves the convergence speed and reliability\cite{li_cyclical_2020}. We limit the training process to a default maximum of 50 epochs and incorporate checkpointing for models with the lowest weighted validation loss, alongside early-stopping. Advanced settings such as batch size, number of epochs, maximum initial learning rate and gamma can be adapted, if required. The best model, in addition with training and evaluation statistics, is exported and downloaded after completion. To ensure comprehensive reporting and documentation, the exported statements include: 1. the number of samples, 2. distribution and corresponding labels, 3. the used base model, 4. the trainings- and evaluation-loss per epoch and 5. the corresponding F1 scores on the evaluation-set. 
\subsection{CardX Foundation Model}
Foundation models, known for their broad applicability and training on diverse data, are increasingly relevant in specialized domains like cardiology. In this paper, we introduce CardX, a state-of-the-art cardiology expert model designed for a variety of downstream ECG analysis tasks. Batch effects due to device variations and calibration are known issues in ECG analysis\cite{plagwitz_necessity_2023}. To ensure robustness against these effects, CardX is trained on a diverse collection of six ECG datasets: PTB-XL\cite{wagner_ptb-xl_2020}, Chapman\cite{Zheng2020,zheng_large_nodate}, Georgia\cite{noauthor_georgia_nodate}, MIMIC-IV-ECG\cite{gow_mimic-iv-ecg_nodate}, SPH\cite{liu_large-scale_2022}, and Deepfake ECGs\cite{thambawita_deepfake_2021} (Table \ref{tab:training-datasets}). Each dataset includes 12-lead ECGs based upon the standardized order of limb leads (I, II, III, aVR, aVL, aVF) and precordial leads (V1-V6), of which all ECGs have been transposed to. For Deepfake-ECGs, the 12 leads have been computed using the provided pre-computed 8-leads dataset and by applying linear transformations via Einthoven and Goldberger equations (III, aVR, aVL, aVF). All ECGs have been unified by downsampling to 100hz, using fast-fourier transformation, and scaled to milliVolt (mV) with a 1000 analog to digital converter (ADC) gain, if applicable. Alongside, a baseline wander removal with a moving median and a window size of 200 ms is applied to reduce variance shifts, according to previous research\cite{lenis_comparison_2017}. For standardization, outlier clipping via quartiles (.01 percentile) without noise filtering has been applied. Internal evaluation has shown that additional normalization reduces performance across datasets. We excluded manually curated folds (9 and 10) from PTB-XL to ensure distinct evaluation and testing. Table \ref{tab:training-datasets} shows the data-, and sex-distribution, along with median age and interquartile range, and whether the dataset is a composite or derived from a single source.
\begin{table*}[ht]
\vspace{1em}
    \centering
    \begin{tabular}{|l|r|rr|rrc|c|}
    \hline
    \multicolumn{1}{|c|}{\multirow{2}{*}{Dataset}} & \multicolumn{1}{c|}{\multirow{2}{*}{\#ECGs}} & \multicolumn{2}{c|}{Age}                               & \multicolumn{3}{c|}{Sex Distribution (\%)}                                 & \multirow{2}{*}{Single Source} \\ \cline{3-7}
    \multicolumn{1}{|c|}{}                         & \multicolumn{1}{c|}{}                        & \multicolumn{1}{c|}{Median} & \multicolumn{1}{c|}{IQR} & \multicolumn{1}{c|}{Male}  & \multicolumn{1}{c|}{Female} & Unknown/Diverse &                                \\ \hline
    PTB-XL*                                        & 17.420                                       & \multicolumn{1}{r|}{62}     & 22                       & \multicolumn{1}{r|}{52}    & \multicolumn{1}{r|}{48}     & -               & Y                              \\ \hline
    Georgia                                        & 10.344                                       & \multicolumn{1}{r|}{62}     & 21                       & \multicolumn{1}{r|}{53.7}  & \multicolumn{1}{r|}{46.3}   & -               & Y                              \\ \hline
    Chapman                                        & 45.152                                       & \multicolumn{1}{r|}{60.5}   & 24                       & \multicolumn{1}{r|}{54}    & \multicolumn{1}{r|}{46}     & -               & N                              \\ \hline
    MIMIC-IV ECG                                   & 800.035                                      & \multicolumn{1}{r|}{66}     & 23                       & \multicolumn{1}{r|}{50.77} & \multicolumn{1}{r|}{48.67}  & 0.56            & Y                              \\ \hline
    SPH                                            & 25.770                                       & \multicolumn{1}{r|}{50}     & 25                       & \multicolumn{1}{r|}{55.36} & \multicolumn{1}{r|}{44.64}  & -               & Y                              \\ \hline
    Deepfake-ECGs                                  & 150.000                                      & \multicolumn{1}{c|}{-}      & \multicolumn{1}{c|}{-}   & \multicolumn{1}{c|}{-}     & \multicolumn{1}{c|}{-}      & -               & Y                              \\ \hline
    \end{tabular}
    \caption{Training sources of the foundation model and corresponding metadata. ${}^*$PTB-XL only stratified folds 1-8.}
    \label{tab:training-datasets}
\end{table*}
\subsection{Architecture}
\begin{figure}[!ht]
\vspace{1em}
    \centering
    \includegraphics[width=0.9\linewidth]{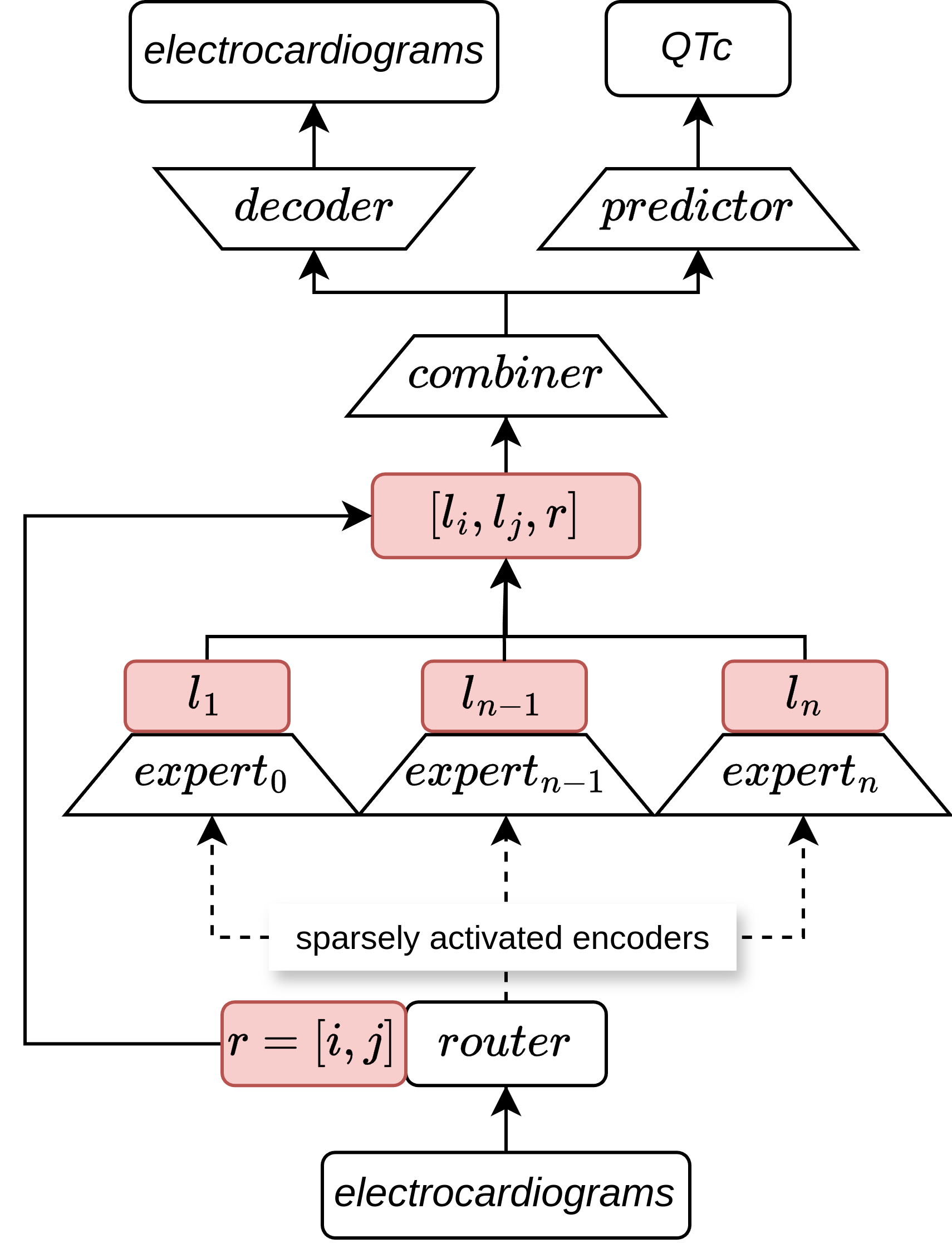}
    \caption{Mixture of Architectures scheme of the electrocardiogram foundation model. The encoders are sparsely activated (dotted line) for each Electrocardiogram, depending on the router.
CardX supports self- and semi-supervised training approaches. The architecture is structured for multi-task learning, capable of addressing multiple objectives: multivariate time-aligned ECG reconstruction, corrected QT-time (QTc) regression, origin dataset prediction, and prevalent feature prediction. For the final foundation model intended for distribution, the reconstruction branch is intentionally removed, and replaced by a linear layer, the classification head. This architectural decision is justified by the pre-training role of reconstruction, its absence of direct clinical utility in downstream tasks, and the resulting advantages in model size reduction and privacy preservation.}
    \label{fig:MoA}
\end{figure}

This work proposes a novel architecture based on the autoencoder framework. The encoder is implemented as a Mixture of Experts (MoE)\cite{jacobs_adaptive_1991} stack, using sparse activation. Expert contributions are combined through the concatenation of the top-k experts, selected by a trainable convolutional router. To facilitate robust training and ensure thorough expert exploration, noisy top-k routing is employed during training, coupled with Gumbel-Softmax\cite{jang_categorical_2017,maddison_concrete_2017} annealing applied to the router logits. Expert activations are further regulated by a combination of load-balance- and route-entropy-loss. Moreover, the architecture introduces a Mixture of Architectures (MoA) concept, utilizing diverse expert architectures within the MoE framework, specifically including XceptionTime, InceptionTime, XResNet\cite{he_deep_2016}, and Transformer\cite{vaswani_attention_2017}. The rationale behind the MoA approach stems from the observed performance heterogeneity across different ECG classes and the data-dependent performance annealing of these architectures\cite{bickmann_post_2023}. The selection of encoder architectures is exposed as a tunable hyperparameter. Finally, to improve the performance of the Transformer architecture, the time Absolute Position Encoding (tAPE), a positional encoding technique specifically developed for time-series data\cite{foumani_improving_2024}, has been incorporated. Additional constraints, such as a variational latent space can be configured. Figure \ref{fig:MoA} shows a schematic of the Mixture of Architectures model.
\subsection{Pretraining}
The model is trained using the AdamW\cite{loshchilov_decoupled_2018} optimizer, an initial learning rate of 1e-3, and gradient clipping with a clip-value of 2. An exponential learning-rate scheduler was applied ($\gamma=0.98$) to facilitate learning rate annealing. For each training dataset, a random 80/20 split was applied to create training and validation sets. Mean Squared Error is employed for the reconstruction task, with an additional Maximum-Mean-Discrepancy\cite{gretton_kernel_2012} loss on the latent space. To mitigate expert dropout, a known issue in sparsely activated MoE architectures, noisy top-k routing (k=2) was integrated with corresponding route-balancing and -entropy loss. To integrate improved efficiency and representational power into our ECG reconstruction model, we incorporate a novel reconstruction task inspired by recent advancements in shallow learning for ECG analysis\cite{plagwitz_rlign_2024}. The Rlign transformation focuses on aligning R-peaks and resampling segments between them, effectively structuring the cyclic nature of ECG signals. This approach yields key advantages for pre-training and model performance. By focusing on the aligned signal, we significantly reduce the necessary temporal embedding, reducing computational load during training and inference. Furthermore, it ensures that the model learns from consistently structured cardiac cycles, explicitly emphasizing diagnostically relevant features like P-waves, QRS complexes, and T-waves within a standardized interval. Finally, it allows for more compressed and informative latent representations. Our approach achieves a 33\% reduction in embedding size (512 variables) compared to a direct competitor (768 variables), while still enabling a latent space that efficiently captures salient ECG features beneficial for downstream tasks like classification or clustering. The data parallel training was conducted on an AMD EPYC 32-Core Processor with three Nvidia A40 40GB GPUs and mixed precision (float16). The utilised batch size is 256 per GPU (768 total).
\subsection{Evaluation}
To thoroughly assess the reliability and performance of ExChanGeAI`s models, we conduct a series of rigorous scenario-based tests. The fine-tuning and prediction capabilities are one of the key features of this work and are therefore mainly evaluated with fine-tuning of models on classification tasks, and evaluation on internal and external datasets using the ExChanGeAI platform where possible.

The evaluation involves training and fine-tuning different (pretrained) architectures across different targets. Specifically, we explore two fine-tuning strategies: 1. Fine-tune only the classification head and 2. Fine-tune all layers. For fine-tuning, we use multiple targets and employ the large open access gold standard PTB-XL dataset. The provided stratified fold 9 is used for training and fold 10 for intra-dataset testing. A single fold is used, to showcase the training of clinical models, with limited data availability. A comparison baseline XceptionTime-model is pre-trained on the folds 1-8 of PTB-XL using this platform. All data is used at a sampling rate of 100 hz. This standardized approach ensures consistent and comparable results. We compare the performance across various levels of difficulty and granularity if applicable. The comparison include: 1. broad diagnostic categories including myocardial infarction (MI), ST/T changes (STTC), conduction disturbances (CD), and hypertrophy (HYP), 2. specific comparisons, such as anterior myocardial infarction (AMI) vs inferior myocardial infarction (IMI), and diverse types of bundle branch blocks including complete left (CLBBB), complete right (CRBBB), and incomplete left (ILBBB). Our tests are executed using the default settings of ExChanGeAI to maintain consistency and integrity across evaluations. As shown by previous research, InceptionTime performs well with less data in comparison to XceptionTime, but falls behind with much data\cite{bickmann_post_2023}. Therefore, we train a baseline model on XceptionTime, as its capability exceeds InceptionTime due to the large amount of data available for the baseline model.

We aim to assess the effectiveness and improvements gained using ExChanGeAI’s fine-tuning capabilities and the possible use of pretrained models, especially in resource constrained environments with very few datapoints. To evaluate model performance, we utilize the F1 score for overall assessment and calculate the average and median for central tendency. Robustness, indicated by lower interquartile range (IQR) and coefficient of variation (CV) values, suggests greater consistency across datasets. Computational scaling is analysed using the number of parameters and FLOPs. These comprehensive evaluations enable us to determine how well models within ExChanGeAI perform under varied conditions, providing insights into its practical application in diverse real-world scenarios. Through this extensive testing framework, we confirm ExChanGeAI's robustness, adaptability, and reliability for ECG analysis across multiple datasets, diagnostic statements and applicability for different use-cases. 

To analyse the applicability of models across sites, we evaluate all models on inter-datasets based on Yang et al., MIMIC-IV-ECG, and Emergency Department Münster (EDMS). This helps to gauge the models generalizability across different datasets. Some classes present slight variations, such as MIMIC and EDMS, which do not have descriptive ECG statements, but general ICD-10 codes. We extract the signals with the corresponding fitting merged superclass. This includes changes such as that bundle branch blocks are only encoded and divided into left- and right-BBB. We evaluate the prediction performance accordingly, counting complete- and incomplete-RBBB as the superclass RBBB. The corresponding used ICD-10-codes, or included statements, are given for each map. Table \ref{tab:test-datasets} shows a comprehensive overview of all extracted targets, number of samples and mapped ICD-10 codes across the different inter-datasets. Note that most datasets are uncurated and reflect real-world implications, in contrast to semi-curated datasets such as PTB-XL. There may be bad quality data, as well as discrepancies between ICD-10 codes and mapped classes. The codes may be based on other Electronic Health Data (EHR), than the ECG, and differences may occur due to indifference between suspected and confirmed diagnosis. An additional case study with manual annotated gold-standard is evaluated on the internal EDMS dataset. This data has been used to train a set of models to predict if revascularization is necessary. It consists of a 240 positive and negative cases each, whereas negative cases are only a subset of the complete annotated dataset. A stratified subset (20\%) of these datapoints are kept as testing data.
\begin{table*}[!ht]
\vspace{1em}
\small
\renewcommand{\arraystretch}{1}
\setlength{\tabcolsep}{4pt}
    \centering
    \begin{tabular}{|l|l|l|ccc}
\hline
Publication & Notes & Target                                            & \multicolumn{3}{c|}{Classes}                                                                \\ \hline
\multirow{4}{*}{Yang et al.} &
  \multirow{4}{*}{\begin{tabular}[c]{@{}l@{}}500hz\\ Based on dataset labels\end{tabular}} &
  \multirow{2}{*}{Superclasses} &
  \multicolumn{1}{c|}{HYP (HEH)} &
  \multicolumn{2}{p{4cm}|}{CD (IAVB, IIAVB1, IIIAVB, BBB, LAFB, NICD)} \\ \cline{4-6} 
            &       &                                                   & \multicolumn{1}{c|}{647}          & \multicolumn{2}{c|}{1974}                               \\ \cline{3-6} 
            &       & \multirow{2}{*}{Bundle branch blocks}             & \multicolumn{1}{c|}{CLBBB}        & \multicolumn{1}{c|}{CRBBB} & \multicolumn{1}{c|}{IRBBB} \\ \cline{4-6} 
            &       &                                                   & \multicolumn{1}{c|}{43}           & \multicolumn{1}{c|}{328}   & \multicolumn{1}{c|}{1051}  \\ \hline
\multirow{6}{*}{MIMIC-IV-ECG} &
  \multirow{6}{*}{\begin{tabular}[c]{@{}l@{}}500hz\\ 200 adu/mV gain\\ Based on ICD-10 codes\end{tabular}} &
  \multirow{2}{*}{Superclasses} &
  \multicolumn{1}{c|}{HYP (I11, I51.7)} &
  \multicolumn{1}{c|}{MI (I21, I22)} &
  \multicolumn{1}{c|}{CD (I44)} \\ \cline{4-6} 
            &       &                                                   & \multicolumn{1}{c|}{500}          & \multicolumn{1}{c|}{500}   & \multicolumn{1}{c|}{500}   \\ \cline{3-6} 
            &       & \multirow{2}{*}{Myocardial infarcts}              & \multicolumn{1}{c|}{AMI (I21.0)}  & \multicolumn{2}{c|}{IMI (I21.1)}                        \\ \cline{4-6} 
            &       &                                                   & \multicolumn{1}{c|}{500}          & \multicolumn{2}{c|}{500}                                \\ \cline{3-6} 
            &       & \multirow{2}{3.1cm}{Bundle branch blocks (variation)}  & \multicolumn{1}{c|}{LBBB (I44.7)} & \multicolumn{2}{c|}{RBBB (I45.1)}                       \\ \cline{4-6} 
            &       &                                                   & \multicolumn{1}{c|}{500}          & \multicolumn{2}{c|}{500}                                \\ \hline
\multirow{8}{*}{EDMS} &
  \multirow{6}{*}{\begin{tabular}[c]{@{}l@{}}100hz\\ Based on ICD-10 codes\end{tabular}} &
  \multirow{2}{*}{Superclasses} &
  \multicolumn{1}{c|}{HYP (I11, I51.7)} &
  \multicolumn{1}{c|}{MI (I21, I22)} &
  \multicolumn{1}{c|}{CD (I44)} \\ \cline{4-5}
            &       &                                                   & \multicolumn{1}{c|}{149}          & \multicolumn{1}{c|}{302}   & \multicolumn{1}{c|}{255}   \\ \cline{3-6} 
            &       & \multirow{2}{*}{Myocardial infarcts}              & \multicolumn{1}{c|}{AMI (I21.0)}  & \multicolumn{2}{c|}{IMI (I21.1)}                        \\ \cline{4-6} 
            &       &                                                   & \multicolumn{1}{c|}{51}           & \multicolumn{2}{c|}{43}                                 \\ \cline{3-6} 
            &       & \multirow{2}{3.1cm}{Bundle branch blocks (variation)} & \multicolumn{1}{c|}{LBBB (I44.7)} & \multicolumn{2}{c|}{RBBB (I45.1)}                       \\ \cline{4-6} 
            &       &                                                   & \multicolumn{1}{c|}{73}           & \multicolumn{2}{c|}{48}                                 \\ \cline{2-6} 
 &
  \multirow{2}{*}{\begin{tabular}[c]{@{}l@{}}500hz\\ Annotated trough cardiologists\end{tabular}} &
  \multirow{2}{*}{Revascularization} &
  \multicolumn{1}{c|}{Yes} &
  \multicolumn{2}{c|}{No} \\ \cline{4-6} 
            &       &                                                   & \multicolumn{1}{c|}{48}           & \multicolumn{2}{c|}{48}                                 \\ \hline
\end{tabular}
    \caption{Dataset and class distribution of the test sets across all defined categories.}
    \label{tab:test-datasets}
    \vspace{1em}
\end{table*}
The only available claimed foundation model ECG-FM is tested on the inter-datasets due to unknown details of training data usage, which is based on Physionet 2021, a collection that includes the PTB-XL dataset as a subset. The model “physionet\_finetuned” is different from the paper results, as these model weights are not accessible. It requires an input of 500hz and 5 seconds samples (first part of the signal) using z-score normalization. The open source weights of the finetuned Physionet model were trained on 26 labels, the same as the pre-trained DSAIL SNU model, which are used as a direct comparison. The labels are mapped to the bundle branch blocks and superclasses where applicable (Table \ref{supp:class-matching}), as no MI classes are included. Classes which are not represented in the Physionet labels are removed for F1 score evaluation. The comparison does not remove any false positives, as this would not be a realistic comparison for a foundation model. To provide a one-to-one comparison, we fine-tuned the pretrained ECG-FM model under the same conditions, such as hyperparameters, optimizer and used PTB-XL fold. We train the final linear classification layer, similar to other models. These models are benchmarked against the inter-datasets as well.

\printbibliography

@article{wagner_ptb-xl_2020,
	title = {{PTB}-{XL}, a large publicly available electrocardiography dataset},
	volume = {7},
	issn = {2052-4463},
	url = {https://www.nature.com/articles/s41597-020-0495-6},
	doi = {10.1038/s41597-020-0495-6},
	
	language = {en},
	number = {1},
	urldate = {2024-03-05},
	journal = {Sci Data},
	author = {Wagner, Patrick and Strodthoff, Nils and Bousseljot, Ralf-Dieter and Kreiseler, Dieter and Lunze, Fatima I. and Samek, Wojciech and Schaeffter, Tobias},
	month = may,
	year = {2020},
	note = {Number: 1},
	pages = {154},
}

@Article{Zheng2020,
author={Zheng, Jianwei
and Chu, Huimin
and Struppa, Daniele
and Zhang, Jianming
and Yacoub, Sir Magdi
and El-Askary, Hesham
and Chang, Anthony
and Ehwerhemuepha, Louis
and Abudayyeh, Islam
and Barrett, Alexander
and Fu, Guohua
and Yao, Hai
and Li, Dongbo
and Guo, Hangyuan
and Rakovski, Cyril},
title={Optimal Multi-Stage Arrhythmia Classification Approach},
journal={Scientific Reports},
year={2020},
month={2},
day={19},
volume={10},
number={1},
pages={2898},
issn={2045-2322},
doi={10.1038/s41598-020-59821-7},
url={https://doi.org/10.1038/s41598-020-59821-7}
}

@ARTICLE{Rauniyar_federated,

  author={Rauniyar, Ashish and Hagos, Desta Haileselassie and Jha, Debesh and Håkegård, Jan Erik and Bagci, Ulas and Rawat, Danda B. and Vlassov, Vladimir},

  journal={IEEE Internet of Things Journal}, 

  title={Federated Learning for Medical Applications: A Taxonomy, Current Trends, Challenges, and Future Research Directions}, 

  year={2024},

  volume={11},

  number={5},

  pages={7374-7398},

  doi={10.1109/JIOT.2023.3329061}}

@article{ismail_fawaz_inceptiontime_2020,
	title = {{InceptionTime}: {Finding} {AlexNet} for time series classification},
	volume = {34},
	issn = {1573-756X},
	shorttitle = {{InceptionTime}},
	url = {https://doi.org/10.1007/s10618-020-00710-y},
	doi = {10.1007/s10618-020-00710-y},
	
	language = {en},
	number = {6},
	urldate = {2024-03-05},
	journal = {Data Min Knowl Disc},
	author = {Ismail Fawaz, Hassan and Lucas, Benjamin and Forestier, Germain and Pelletier, Charlotte and Schmidt, Daniel F. and Weber, Jonathan and Webb, Geoffrey I. and Idoumghar, Lhassane and Muller, Pierre-Alain and Petitjean, François},
	month = nov,
	year = {2020},
	note = {Number: 6},
	pages = {1936--1962},
}

@article{goldberger_physiobank_2000,
	title = {{PhysioBank}, {PhysioToolkit}, and {PhysioNet}},
	volume = {101},
	url = {https://www.ahajournals.org/doi/10.1161/01.cir.101.23.e215},
	doi = {10.1161/01.CIR.101.23.e215},
	
	number = {23},
	urldate = {2024-03-05},
	journal = {Circulation},
	author = {Goldberger, Ary L. and Amaral, Luis A. N. and Glass, Leon and Hausdorff, Jeffrey M. and Ivanov, Plamen Ch. and Mark, Roger G. and Mietus, Joseph E. and Moody, George B. and Peng, Chung-Kang and Stanley, H. Eugene},
	month = jun,
	year = {2000},
	note = {Number: 23
Publisher: American Heart Association},
	pages = {e215--e220},
}

@article{bickmann_post_2023,
	title = {Post {Hoc} {Sample} {Size} {Estimation} for {Deep} {Learning} {Architectures} for {ECG}-{Classification}},
	volume = {302},
	copyright = {cc by-nc},
	issn = {1879-8365},
	url = {https://doi.org/10.3233/SHTI230099},
	doi = {10.3233/shti230099},
	
	language = {eng},
	urldate = {2024-03-05},
	journal = {Stud Health Technol Inform},
	author = {Bickmann, Lucas and Plagwitz, Lucas and Varghese, Julian},
	month = may,
	year = {2023},
	pmid = {37203643},
	pages = {182--186},
}

@article{plagwitz_necessity_2023,
	title = {The {Necessity} of {Multiple} {Data} {Sources} for {ECG}-{Based} {Machine} {Learning} {Models}},
	volume = {302},
	copyright = {cc by-nc},
	issn = {1879-8365},
	url = {https://doi.org/10.3233/SHTI230059},
	doi = {10.3233/shti230059},
	
	language = {eng},
	urldate = {2024-03-05},
	journal = {Stud Health Technol Inform},
	author = {Plagwitz, Lucas and Vogelsang, Tobias and Doldi, Florian and Bickmann, Lucas and Fujarski, Michael and Eckardt, Lars and Varghese, Julian},
	month = may,
	year = {2023},
	pmid = {37203604},
	pages = {33--37},
}

@article{avetisyan_deep_2024,
	title = {Deep neural networks generalization and fine-tuning for 12-lead {ECG} classification},
	volume = {93},
	issn = {1746-8094},
	url = {https://www.sciencedirect.com/science/article/pii/S1746809424002180},
	doi = {10.1016/j.bspc.2024.106160},
	
	urldate = {2024-07-30},
	journal = {Biomedical Signal Processing and Control},
	author = {Avetisyan, Aram and Tigranyan, Shahane and Asatryan, Ariana and Mashkova, Olga and Skorik, Sergey and Ananev, Vladislav and Markin, Yury},
	month = jul,
	year = {2024},
	pages = {106160},
}

@article{kaplan_berkaya_survey_2018,
	title = {A survey on {ECG} analysis},
	volume = {43},
	issn = {1746-8094},
	url = {https://www.sciencedirect.com/science/article/pii/S1746809418300636},
	doi = {10.1016/j.bspc.2018.03.003},
	urldate = {2024-07-30},
	journal = {Biomedical Signal Processing and Control},
	author = {Kaplan Berkaya, Selcan and Uysal, Alper Kursat and Sora Gunal, Efnan and Ergin, Semih and Gunal, Serkan and Gulmezoglu, M. Bilginer},
	month = may,
	year = {2018},
	pages = {216--235},
}

@article{qin_lightweight_2024,
	title = {A lightweight {SelfONN} model for general {ECG} classification with pretraining},
	volume = {89},
	issn = {1746-8094},
	url = {https://www.sciencedirect.com/science/article/pii/S1746809423012132},
	doi = {10.1016/j.bspc.2023.105780},
	urldate = {2024-07-30},
	journal = {Biomedical Signal Processing and Control},
	author = {Qin, Keke and Huang, Wu and Zhang, Tao and Zhang, Hengyuan and Cheng, Xiangrong},
	month = mar,
	year = {2024},
	pages = {105780},
}

@article{weimann_transfer_2021,
	title = {Transfer learning for {ECG} classification},
	volume = {11},
	copyright = {2021 The Author(s)},
	issn = {2045-2322},
	url = {https://www.nature.com/articles/s41598-021-84374-8},
	doi = {10.1038/s41598-021-84374-8},
	language = {en},
	number = {1},
	urldate = {2024-07-30},
	journal = {Sci Rep},
	author = {Weimann, Kuba and Conrad, Tim O. F.},
	month = mar,
	year = {2021},
	note = {Number: 1
Publisher: Nature Publishing Group},
	pages = {5251},
}

@article{ebrahimi_review_2020,
	title = {A review on deep learning methods for {ECG} arrhythmia classification},
	volume = {7},
	issn = {2590-1885},
	url = {https://www.sciencedirect.com/science/article/pii/S2590188520300123},
	doi = {10.1016/j.eswax.2020.100033},
	urldate = {2024-07-30},
	journal = {Expert Systems with Applications: X},
	author = {Ebrahimi, Zahra and Loni, Mohammad and Daneshtalab, Masoud and Gharehbaghi, Arash},
	month = sep,
	year = {2020},
	pages = {100033},
}

@article{makowski_neurokit2_2021,
	title = {{NeuroKit2}: {A} {Python} toolbox for neurophysiological signal processing},
	volume = {53},
	issn = {1554-3528},
	shorttitle = {{NeuroKit2}},
	url = {https://doi.org/10.3758/s13428-020-01516-y},
	doi = {10.3758/s13428-020-01516-y},
	language = {en},
	number = {4},
	urldate = {2024-07-30},
	journal = {Behav Res},
	author = {Makowski, Dominique and Pham, Tam and Lau, Zen J. and Brammer, Jan C. and Lespinasse, François and Pham, Hung and Schölzel, Christopher and Chen, S. H. Annabel},
	month = aug,
	year = {2021},
	note = {Number: 4},
	pages = {1689--1696},
}

@inproceedings{reyna_will_2021,
	address = {Brno, Czech Republic},
	title = {Will {Two} {Do}? {Varying} {Dimensions} in {Electrocardiography}: {The} {PhysioNet}/{Computing} in {Cardiology} {Challenge} 2021},
	copyright = {https://doi.org/10.15223/policy-029},
	isbn = {978-1-6654-7916-5},
	shorttitle = {Will {Two} {Do}?},
	url = {https://ieeexplore.ieee.org/document/9662687/},
	doi = {10.23919/CinC53138.2021.9662687},
	language = {en},
	urldate = {2024-07-30},
	booktitle = {2021 {Computing} in {Cardiology} ({CinC})},
	publisher = {IEEE},
	author = {Reyna, Matthew A and Sadr, Nadi and Alday, Erick A Perez and Gu, Annie and Shah, Amit J and Robichaux, Chad and Rad, Ali Bahrami and Elola, Andoni and Seyedi, Salman and Ansari, Sardar and Ghanbari, Hamid and Li, Qiao and Sharma, Ashish and Clifford, Gari D},
	month = sep,
	year = {2021},
	pages = {1--4},
}

@article{reyna_issues_2022,
	title = {Issues in the automated classification of multilead ecgs using heterogeneous labels and populations},
	volume = {43},
	issn = {0967-3334, 1361-6579},
	url = {https://iopscience.iop.org/article/10.1088/1361-6579/ac79fd},
	doi = {10.1088/1361-6579/ac79fd},
	language = {en},
	number = {8},
	urldate = {2024-07-30},
	journal = {Physiol. Meas.},
	author = {Reyna, Matthew A and Sadr, Nadi and Perez Alday, Erick A and Gu, Annie and Shah, Amit J and Robichaux, Chad and Bahrami Rad, Ali and Elola, Andoni and Seyedi, Salman and Ansari, Sardar and Ghanbari, Hamid and Li, Qiao and Sharma, Ashish and Clifford, Gari D},
	month = aug,
	year = {2022},
	note = {Number: 8},
	pages = {084001},
}

@inproceedings{han_towards_2021,
	address = {Brno, Czech Republic},
	title = {Towards {High} {Generalization} {Performance} on {Electrocardiogram} {Classification}},
	copyright = {https://doi.org/10.15223/policy-029},
	isbn = {978-1-6654-7916-5},
	url = {https://ieeexplore.ieee.org/document/9662737/},
	doi = {10.23919/CinC53138.2021.9662737},
	language = {en},
	urldate = {2024-07-30},
	booktitle = {2021 {Computing} in {Cardiology} ({CinC})},
	publisher = {IEEE},
	author = {Han, Hyeongrok and Park, Seongjae and Min, Seonwoo and Choi, Hyun-Soo and Kim, Eunji and Kim, Hyunki and Park, Sangha and Kim, Jinkook and Park, Junsang and An, Junho and Lee, Kwanglo and Jeong, Wonsun and Chon, Sangil and Ha, Kwonwoo and Han, Myungkyu and Yoon, Sungroh},
	month = sep,
	year = {2021},
	pages = {1--4},
}

@inproceedings{rahimian_xceptiontime_2020,
	title = {{XceptionTime}: {Independent} {Time}-{Window} {Xceptiontime} {Architecture} for {Hand} {Gesture} {Classification}},
	shorttitle = {{XceptionTime}},
	url = {https://ieeexplore.ieee.org/document/9054586},
	doi = {10.1109/ICASSP40776.2020.9054586},
	urldate = {2024-07-30},
	booktitle = {{ICASSP} 2020 - 2020 {IEEE} {International} {Conference} on {Acoustics}, {Speech} and {Signal} {Processing} ({ICASSP})},
	author = {Rahimian, Elahe and Zabihi, Soheil and Atashzar, Seyed Farokh and Asif, Amir and Mohammadi, Arash},
	month = may,
	year = {2020},
	note = {ISSN: 2379-190X},
	pages = {1304--1308},
}

@misc{developers_onnx2torch_2021,
	title = {onnx2torch},
	copyright = {Apache-2.0},
	url = {https://github.com/ENOT-AutoDL/onnx2torch},
	urldate = {2025-02-26},
	author = {developers, ENOT and Kalgin, Igor and Yanchenko, Arseny and Ivanov, Pyoter and Goncharenko, Alexander},
	month = dec,
	year = {2021},
}

@inproceedings{ansel_pytorch_2024,
	address = {New York, NY, USA},
	series = {{ASPLOS} '24},
	title = {{PyTorch} 2: {Faster} {Machine} {Learning} {Through} {Dynamic} {Python} {Bytecode} {Transformation} and {Graph} {Compilation}},
	volume = {2},
	isbn = {979-8-4007-0385-0},
	shorttitle = {{PyTorch} 2},
	url = {https://dl.acm.org/doi/10.1145/3620665.3640366},
	doi = {10.1145/3620665.3640366},
	urldate = {2024-08-20},
	booktitle = {Proceedings of the 29th {ACM} {International} {Conference} on {Architectural} {Support} for {Programming} {Languages} and {Operating} {Systems}, {Volume} 2},
	publisher = {Association for Computing Machinery},
	author = {Ansel, Jason and Yang, Edward and He, Horace and Gimelshein, Natalia and Jain, Animesh and Voznesensky, Michael and Bao, Bin and Bell, Peter and Berard, David and Burovski, Evgeni and Chauhan, Geeta and Chourdia, Anjali and Constable, Will and Desmaison, Alban and DeVito, Zachary and Ellison, Elias and Feng, Will and Gong, Jiong and Gschwind, Michael and Hirsh, Brian and Huang, Sherlock and Kalambarkar, Kshiteej and Kirsch, Laurent and Lazos, Michael and Lezcano, Mario and Liang, Yanbo and Liang, Jason and Lu, Yinghai and Luk, C. K. and Maher, Bert and Pan, Yunjie and Puhrsch, Christian and Reso, Matthias and Saroufim, Mark and Siraichi, Marcos Yukio and Suk, Helen and Zhang, Shunting and Suo, Michael and Tillet, Phil and Zhao, Xu and Wang, Eikan and Zhou, Keren and Zou, Richard and Wang, Xiaodong and Mathews, Ajit and Wen, William and Chanan, Gregory and Wu, Peng and Chintala, Soumith},
	month = apr,
	year = {2024},
	pages = {929--947},
}

@article{lai_practical_2023,
	title = {Practical intelligent diagnostic algorithm for wearable 12-lead {ECG} via self-supervised learning on large-scale dataset},
	volume = {14},
	copyright = {2023 The Author(s)},
	issn = {2041-1723},
	url = {https://www.nature.com/articles/s41467-023-39472-8},
	doi = {10.1038/s41467-023-39472-8},
	language = {en},
	number = {1},
	urldate = {2024-08-20},
	journal = {Nat Commun},
	author = {Lai, Jiewei and Tan, Huixin and Wang, Jinliang and Ji, Lei and Guo, Jun and Han, Baoshi and Shi, Yajun and Feng, Qianjin and Yang, Wei},
	month = jun,
	year = {2023},
	note = {Number: 1
Publisher: Nature Publishing Group},
	pages = {3741},
}

@misc{yang_offline_2023,
	title = {Offline {Test} {Set} of {ECG} {Multi}-label {Classfication}},
	url = {https://www.scidb.cn/en/detail?dataSetId=58c4a92d5a01414390a78160d335380d},
	doi = {10.57760/sciencedb.07677},
	publisher = {Science Data Bank},
	author = {Yang, Wei and Feng, Qianjin},
	year = {2023},
}

@article{rahman_impact_2024,
	title = {Impact of {Artificial} {Intelligence} ({AI}) {Technology} in {Healthcare} {Sector}: {A} {Critical} {Evaluation} of {Both} {Sides} of the {Coin}},
	volume = {17},
	issn = {2632-010X},
	shorttitle = {Impact of {Artificial} {Intelligence} ({AI}) {Technology} in {Healthcare} {Sector}},
	url = {https://www.ncbi.nlm.nih.gov/pmc/articles/PMC10804900/},
	doi = {10.1177/2632010X241226887},
	urldate = {2024-12-04},
	journal = {Clin Pathol},
	author = {Rahman, Md. Ashrafur and Victoros, Evangelos and Ernest, Julianne and Davis, Rob and Shanjana, Yeasna and Islam, Md. Rabiul},
	month = jan,
	year = {2024},
	pmid = {38264676},
	pmcid = {PMC10804900},
	pages = {2632010X241226887},
}

@article{zhang_ecg_2024,
	title = {{ECG} autoencoder based on low-rank attention},
	volume = {14},
	copyright = {2024 The Author(s)},
	issn = {2045-2322},
	url = {https://www.nature.com/articles/s41598-024-63378-0},
	doi = {10.1038/s41598-024-63378-0},
	language = {en},
	number = {1},
	urldate = {2024-12-04},
	journal = {Sci Rep},
	author = {Zhang, Shilin and Fang, Yixian and Ren, Yuwei},
	month = jun,
	year = {2024},
	note = {Publisher: Nature Publishing Group},
	pages = {12823},
}

@article{roy_ecg-net_2023,
	title = {{ECG}-{NET}: {A} deep {LSTM} autoencoder for detecting anomalous {ECG}},
	volume = {124},
	issn = {0952-1976},
	shorttitle = {{ECG}-{NET}},
	url = {https://www.sciencedirect.com/science/article/pii/S0952197623006681},
	doi = {10.1016/j.engappai.2023.106484},
	urldate = {2024-12-04},
	journal = {Engineering Applications of Artificial Intelligence},
	author = {Roy, Moumita and Majumder, Sukanta and Halder, Anindya and Biswas, Utpal},
	month = sep,
	year = {2023},
	pages = {106484},
}

@article{singh_attention-based_2022,
	title = {Attention-{Based} {Convolutional} {Denoising} {Autoencoder} for {Two}-{Lead} {ECG} {Denoising} and {Arrhythmia} {Classification}},
	volume = {71},
	issn = {1557-9662},
	url = {https://ieeexplore.ieee.org/document/9853604},
	doi = {10.1109/TIM.2022.3197757},
	urldate = {2024-12-04},
	journal = {IEEE Transactions on Instrumentation and Measurement},
	author = {Singh, Prateek and Sharma, Ambalika},
	year = {2022},
	note = {Conference Name: IEEE Transactions on Instrumentation and Measurement},
	pages = {1--10},
}

@article{mathew_foundation_2024,
	title = {Foundation models for cardiovascular disease detection via biosignals from digital stethoscopes},
	volume = {1},
	copyright = {2024 The Author(s)},
	issn = {2948-2836},
	url = {https://www.nature.com/articles/s44325-024-00027-5},
	doi = {10.1038/s44325-024-00027-5},
	language = {en},
	number = {1},
	urldate = {2024-12-04},
	journal = {npj Cardiovasc Health},
	author = {Mathew, George and Barbosa, Daniel and Prince, John and Venkatraman, Subramaniam},
	month = oct,
	year = {2024},
	note = {Publisher: Nature Publishing Group},
	pages = {1--13},
}

@misc{salimi_exploring_2023,
	title = {Exploring {Best} {Practices} for {ECG} {Signal} {Processing} in {Machine} {Learning}},
	url = {http://arxiv.org/abs/2311.04229},
	doi = {10.48550/arXiv.2311.04229},
	urldate = {2024-12-04},
	publisher = {arXiv},
	author = {Salimi, Amir and Kalmady, Sunil Vasu and Hindle, Abram and Zaiane, Osmar and Kaul, Padma},
	month = nov,
	year = {2023},
	note = {arXiv:2311.04229 [eess]},
}

@misc{mckeen_ecg-fm_2024,
	title = {{ECG}-{FM}: {An} {Open} {Electrocardiogram} {Foundation} {Model}},
	shorttitle = {{ECG}-{FM}},
	url = {http://arxiv.org/abs/2408.05178},
	doi = {10.48550/arXiv.2408.05178},
	urldate = {2024-12-04},
	publisher = {arXiv},
	author = {McKeen, Kaden and Oliva, Laura and Masood, Sameer and Toma, Augustin and Rubin, Barry and Wang, Bo},
	month = aug,
	year = {2024},
	note = {arXiv:2408.05178 [cs]
version: 1},
}

@misc{li_electrocardiogram_2024,
	title = {An {Electrocardiogram} {Foundation} {Model} {Built} on over 10 {Million} {Recordings} with {External} {Evaluation} across {Multiple} {Domains}},
	url = {http://arxiv.org/abs/2410.04133},
	doi = {10.48550/arXiv.2410.04133},
	urldate = {2024-12-04},
	publisher = {arXiv},
	author = {Li, Jun and Aguirre, Aaron and Moura, Junior and Liu, Che and Zhong, Lanhai and Sun, Chenxi and Clifford, Gari and Westover, Brandon and Hong, Shenda},
	month = oct,
	year = {2024},
	note = {arXiv:2410.04133 [cs]},
}

@article{widder_why_2024,
	title = {Why ‘open’ {AI} systems are actually closed, and why this matters},
	volume = {635},
	copyright = {2024 Springer Nature Limited},
	issn = {1476-4687},
	url = {https://www.nature.com/articles/s41586-024-08141-1},
	doi = {10.1038/s41586-024-08141-1},
	language = {en},
	number = {8040},
	urldate = {2025-01-24},
	journal = {Nature},
	author = {Widder, David Gray and Whittaker, Meredith and West, Sarah Myers},
	month = nov,
	year = {2024},
	note = {Publisher: Nature Publishing Group},
	pages = {827--833},
}

@article{sau_artificial_2024,
	title = {Artificial intelligence-enabled electrocardiogram for mortality and cardiovascular risk estimation: a model development and validation study},
	volume = {6},
	issn = {2589-7500},
	shorttitle = {Artificial intelligence-enabled electrocardiogram for mortality and cardiovascular risk estimation},
	url = {https://www.sciencedirect.com/science/article/pii/S2589750024001729},
	doi = {10.1016/S2589-7500(24)00172-9},
	number = {11},
	urldate = {2025-02-26},
	journal = {The Lancet Digital Health},
	author = {Sau, Arunashis and Pastika, Libor and Sieliwonczyk, Ewa and Patlatzoglou, Konstantinos and Ribeiro, Antônio H and McGurk, Kathryn A and Zeidaabadi, Boroumand and Zhang, Henry and Macierzanka, Krzysztof and Mandic, Danilo and Sabino, Ester and Giatti, Luana and Barreto, Sandhi M and Camelo, Lidyane do Valle and Tzoulaki, Ioanna and O'Regan, Declan P and Peters, Nicholas S and Ware, James S and Ribeiro, Antonio Luiz P and Kramer, Daniel B and Waks, Jonathan W and Ng, Fu Siong},
	month = nov,
	year = {2024},
	pages = {e791--e802},
}

@article{adedinsewo_artificial_2024,
	title = {Artificial intelligence guided screening for cardiomyopathies in an obstetric population: a pragmatic randomized clinical trial},
	volume = {30},
	copyright = {2024 The Author(s)},
	issn = {1546-170X},
	shorttitle = {Artificial intelligence guided screening for cardiomyopathies in an obstetric population},
	url = {https://www.nature.com/articles/s41591-024-03243-9},
	doi = {10.1038/s41591-024-03243-9},
	language = {en},
	number = {10},
	urldate = {2025-02-26},
	journal = {Nat Med},
	author = {Adedinsewo, Demilade A. and Morales-Lara, Andrea Carolina and Afolabi, Bosede B. and Kushimo, Oyewole A. and Mbakwem, Amam C. and Ibiyemi, Kehinde F. and Ogunmodede, James Ayodele and Raji, Hadijat Olaide and Ringim, Sadiq H. and Habib, Abdullahi A. and Hamza, Sabiu M. and Ogah, Okechukwu S. and Obajimi, Gbolahan and Saanu, Olugbenga Oluseun and Jagun, Olusoji E. and Inofomoh, Francisca O. and Adeolu, Temitope and Karaye, Kamilu M. and Gaya, Sule A. and Alfa, Isiaka and Yohanna, Cynthia and Venkatachalam, K. L. and Dugan, Jennifer and Yao, Xiaoxi and Sledge, Hanna J. and Johnson, Patrick W. and Wieczorek, Mikolaj A. and Attia, Zachi I. and Phillips, Sabrina D. and Yamani, Mohamad H. and Tobah, Yvonne Butler and Rose, Carl H. and Sharpe, Emily E. and Lopez-Jimenez, Francisco and Friedman, Paul A. and Noseworthy, Peter A. and Carter, Rickey E.},
	month = oct,
	year = {2024},
	note = {Publisher: Nature Publishing Group},
	pages = {2897--2906},
}

@article{lin_ai-enabled_2024,
	title = {{AI}-enabled electrocardiography alert intervention and all-cause mortality: a pragmatic randomized clinical trial},
	volume = {30},
	copyright = {2024 The Author(s), under exclusive licence to Springer Nature America, Inc.},
	issn = {1546-170X},
	shorttitle = {{AI}-enabled electrocardiography alert intervention and all-cause mortality},
	url = {https://www.nature.com/articles/s41591-024-02961-4},
	doi = {10.1038/s41591-024-02961-4},
	language = {en},
	number = {5},
	urldate = {2025-02-26},
	journal = {Nat Med},
	author = {Lin, Chin-Sheng and Liu, Wei-Ting and Tsai, Dung-Jang and Lou, Yu-Sheng and Chang, Chiao-Hsiang and Lee, Chiao-Chin and Fang, Wen-Hui and Wang, Chih-Chia and Chen, Yen-Yuan and Lin, Wei-Shiang and Cheng, Cheng-Chung and Lee, Chia-Cheng and Wang, Chih-Hung and Tsai, Chien-Sung and Lin, Shih-Hua and Lin, Chin},
	month = may,
	year = {2024},
	note = {Publisher: Nature Publishing Group},
	pages = {1461--1470},
}

@article{polevikov_advancing_2023,
	title = {Advancing {AI} in healthcare: {A} comprehensive review of best practices},
	volume = {548},
	issn = {0009-8981},
	shorttitle = {Advancing {AI} in healthcare},
	url = {https://www.sciencedirect.com/science/article/pii/S0009898123003212},
	doi = {10.1016/j.cca.2023.117519},
	urldate = {2025-02-26},
	journal = {Clinica Chimica Acta},
	author = {Polevikov, Sergei},
	month = aug,
	year = {2023},
	pages = {117519},
}

@misc{han_foundation_2024,
	title = {Foundation {Models} in {Electrocardiogram}: {A} {Review}},
	shorttitle = {Foundation {Models} in {Electrocardiogram}},
	url = {http://arxiv.org/abs/2410.19877},
	doi = {10.48550/arXiv.2410.19877},
	urldate = {2025-03-06},
	publisher = {arXiv},
	author = {Han, Yu and Liu, Xiaofeng and Zhang, Xiang and Ding, Cheng},
	month = nov,
	year = {2024},
	note = {arXiv:2410.19877 [eess]},
}

@misc{wang_anyecg_2025,
	title = {{AnyECG}: {Foundational} {Models} for {Multitask} {Cardiac} {Analysis} in {Real}-{World} {Settings}},
	shorttitle = {{AnyECG}},
	url = {http://arxiv.org/abs/2411.17711},
	doi = {10.48550/arXiv.2411.17711},
	urldate = {2025-03-06},
	publisher = {arXiv},
	author = {Wang, Yue and Cao, Xu and Hu, Yaojun and Ying, Haochao and Xu, Hongxia and Wu, Ruijia and Rehg, James Matthew and Sun, Jimeng and Wu, Jian and Chen, Jintai},
	month = mar,
	year = {2025},
	note = {arXiv:2411.17711 [eess]},
}

@misc{plagwitz_rlign_2024,
	title = {The {Rlign} {Algorithm} for {Enhanced} {Electrocardiogram} {Analysis} through {R}-{Peak} {Alignment} for {Explainable} {Classification} and {Clustering}},
	url = {http://arxiv.org/abs/2407.15555},
	doi = {10.48550/arXiv.2407.15555},

	urldate = {2025-03-06},
	publisher = {arXiv},
	author = {Plagwitz, Lucas and Bickmann, Lucas and Fujarski, Michael and Brenner, Alexander and Gobalakrishnan, Warnes and Eckardt, Lars and Büscher, Antonius and Varghese, Julian},
	month = aug,
	year = {2024},
	note = {arXiv:2407.15555 [eess]},
}

@inproceedings{loshchilov_decoupled_2018,
	title = {Decoupled {Weight} {Decay} {Regularization}},
	url = {https://openreview.net/forum?id=Bkg6RiCqY7},
	
	language = {en},
	urldate = {2025-03-06},
	author = {Loshchilov, Ilya and Hutter, Frank},
	month = sep,
	year = {2018},
}

@inproceedings{li_cyclical_2020,
	title = {A {Cyclical} {Learning} {Rate} {Method} in {Deep} {Learning} {Training}},
	url = {https://ieeexplore.ieee.org/abstract/document/9232482},
	doi = {10.1109/CITS49457.2020.9232482},
	
	urldate = {2025-03-06},
	booktitle = {2020 {International} {Conference} on {Computer}, {Information} and {Telecommunication} {Systems} ({CITS})},
	author = {Li, Jiaqi and Yang, Xiaodong},
	month = oct,
	year = {2020},
	pages = {1--5},
}

@misc{zheng_large_nodate,
	title = {A large scale 12-lead electrocardiogram database for arrhythmia study},
	url = {https://physionet.org/content/ecg-arrhythmia/1.0.0/},
	doi = {10.13026/WGEX-ER52},
	
	urldate = {2025-03-06},
	publisher = {PhysioNet},
	author = {Zheng, Jianwei and Guo, Hangyuan and Chu, Huimin},
}

@misc{gow_mimic-iv-ecg_nodate,
	title = {{MIMIC}-{IV}-{ECG}: {Diagnostic} {Electrocardiogram} {Matched} {Subset}},
	shorttitle = {{MIMIC}-{IV}-{ECG}},
	url = {https://physionet.org/content/mimic-iv-ecg/1.0/},
	doi = {10.13026/4NQG-SB35},
	
	urldate = {2025-03-06},
	publisher = {PhysioNet},
	author = {Gow, Brian and Pollard, Tom and Nathanson, Larry A and Johnson, Alistair and Moody, Benjamin and Fernandes, Chrystinne and Greenbaum, Nathaniel and Waks, Jonathan W and Eslami, Parastou and Carbonati, Tanner and Chaudhari, Ashish and Herbst, Elizabeth and Moukheiber, Dana and Berkowitz, Seth and Mark, Roger and Horng, Steven},
}

@article{liu_large-scale_2022,
	title = {A large-scale multi-label 12-lead electrocardiogram database with standardized diagnostic statements},
	volume = {9},
	copyright = {2022 The Author(s)},
	issn = {2052-4463},
	url = {https://www.nature.com/articles/s41597-022-01403-5},
	doi = {10.1038/s41597-022-01403-5},
	
	language = {en},
	number = {1},
	urldate = {2025-03-06},
	journal = {Sci Data},
	author = {Liu, Hui and Chen, Dan and Chen, Da and Zhang, Xiyu and Li, Huijie and Bian, Lipan and Shu, Minglei and Wang, Yinglong},
	month = jun,
	year = {2022},
	note = {Publisher: Nature Publishing Group},
	pages = {272},
}

@article{thambawita_deepfake_2021,
	title = {{DeepFake} electrocardiograms using generative adversarial networks are the beginning of the end for privacy issues in medicine},
	volume = {11},
	copyright = {2021 The Author(s)},
	issn = {2045-2322},
	url = {https://www.nature.com/articles/s41598-021-01295-2},
	doi = {10.1038/s41598-021-01295-2},
	
	language = {en},
	number = {1},
	urldate = {2025-03-06},
	journal = {Sci Rep},
	author = {Thambawita, Vajira and Isaksen, Jonas L. and Hicks, Steven A. and Ghouse, Jonas and Ahlberg, Gustav and Linneberg, Allan and Grarup, Niels and Ellervik, Christina and Olesen, Morten Salling and Hansen, Torben and Graff, Claus and Holstein-Rathlou, Niels-Henrik and Strümke, Inga and Hammer, Hugo L. and Maleckar, Mary M. and Halvorsen, Pål and Riegler, Michael A. and Kanters, Jørgen K.},
	month = nov,
	year = {2021},
	note = {Publisher: Nature Publishing Group},
	pages = {21896},
}

@article{lenis_comparison_2017,
	title = {Comparison of {Baseline} {Wander} {Removal} {Techniques} considering the {Preservation} of {ST} {Changes} in the {Ischemic} {ECG}: {A} {Simulation} {Study}},
	volume = {2017},
	copyright = {Copyright © 2017 Gustavo Lenis et al.},
	issn = {1748-6718},
	shorttitle = {Comparison of {Baseline} {Wander} {Removal} {Techniques} considering the {Preservation} of {ST} {Changes} in the {Ischemic} {ECG}},
	url = {onlinelibrary.wiley.com/doi/abs/10.1155/2017/9295029},
	doi = {10.1155/2017/9295029},
	
	language = {en},
	number = {1},
	urldate = {2025-03-06},
	journal = {Computational and Mathematical Methods in Medicine},
	author = {Lenis, Gustavo and Pilia, Nicolas and Loewe, Axel and Schulze, Walther H. W. and Dössel, Olaf},
	year = {2017},
	pages = {9295029},
}

@article{foumani_improving_2024,
	title = {Improving position encoding of transformers for multivariate time series classification},
	volume = {38},
	issn = {1573-756X},
	url = {https://doi.org/10.1007/s10618-023-00948-2},
	doi = {10.1007/s10618-023-00948-2},
	
	language = {en},
	number = {1},
	urldate = {2025-03-06},
	journal = {Data Min Knowl Disc},
	author = {Foumani, Navid Mohammadi and Tan, Chang Wei and Webb, Geoffrey I. and Salehi, Mahsa},
	month = jan,
	year = {2024},
	pages = {22--48},
}

@article{jacobs_adaptive_1991,
	title = {Adaptive {Mixtures} of {Local} {Experts}},
	volume = {3},
	issn = {0899-7667},
	url = {https://ieeexplore.ieee.org/abstract/document/6797059},
	doi = {10.1162/neco.1991.3.1.79},
	
	number = {1},
	urldate = {2025-03-06},
	journal = {Neural Computation},
	author = {Jacobs, Robert A. and Jordan, Michael I. and Nowlan, Steven J. and Hinton, Geoffrey E.},
	month = mar,
	year = {1991},
	note = {Conference Name: Neural Computation},
	pages = {79--87},
}

@inproceedings{he_deep_2016,
	title = {Deep {Residual} {Learning} for {Image} {Recognition}},
	url = {https://ieeexplore.ieee.org/document/7780459},
	doi = {10.1109/CVPR.2016.90},
	
	urldate = {2025-03-06},
	booktitle = {2016 {IEEE} {Conference} on {Computer} {Vision} and {Pattern} {Recognition} ({CVPR})},
	author = {He, Kaiming and Zhang, Xiangyu and Ren, Shaoqing and Sun, Jian},
	month = jun,
	year = {2016},
	note = {ISSN: 1063-6919},
	pages = {770--778},
}

@inproceedings{vaswani_attention_2017,
	address = {Red Hook, NY, USA},
	series = {{NIPS}'17},
	title = {Attention is all you need},
	isbn = {978-1-5108-6096-4},
	
	urldate = {2025-03-05},
	booktitle = {Proceedings of the 31st {International} {Conference} on {Neural} {Information} {Processing} {Systems}},
	publisher = {Curran Associates Inc.},
	author = {Vaswani, Ashish and Shazeer, Noam and Parmar, Niki and Uszkoreit, Jakob and Jones, Llion and Gomez, Aidan N. and Kaiser, Łukasz and Polosukhin, Illia},
	month = dec,
	year = {2017},
	pages = {6000--6010},
}

@inproceedings{maddison_concrete_2017,
	title = {The {Concrete} {Distribution}: {A} {Continuous} {Relaxation} of {Discrete} {Random} {Variables}},
	shorttitle = {The {Concrete} {Distribution}},
	url = {https://openreview.net/forum?id=S1jE5L5gl},
	
	language = {en},
	urldate = {2025-03-06},
	author = {Maddison, Chris J. and Mnih, Andriy and Teh, Yee Whye},
	month = feb,
	year = {2017},
}

@inproceedings{jang_categorical_2017,
	title = {Categorical {Reparameterization} with {Gumbel}-{Softmax}},
	url = {https://openreview.net/forum?id=rkE3y85ee},
	
	language = {en},
	urldate = {2025-03-06},
	author = {Jang, Eric and Gu, Shixiang and Poole, Ben},
	month = feb,
	year = {2017},
}

@article{gretton_kernel_2012,
	title = {A kernel two-sample test},
	volume = {13},
	issn = {1532-4435},
	
	number = {null},
	journal = {J. Mach. Learn. Res.},
	author = {Gretton, Arthur and Borgwardt, Karsten M. and Rasch, Malte J. and Schölkopf, Bernhard and Smola, Alexander},
	month = mar,
	year = {2012},
	pages = {723--773},
}

@misc{noauthor_georgia_nodate,
	title = {Georgia 12-{Lead} {ECG} {Challenge} {Database}},
	url = {https://www.kaggle.com/datasets/bjoernjostein/georgia-12lead-ecg-challenge-database},
	language = {en},
	urldate = {2025-03-06},
}

@article{martinez-selles_current_2023,
	title = {Current and {Future} {Use} of {Artificial} {Intelligence} in {Electrocardiography}},
	volume = {10},
	issn = {2308-3425},
	url = {https://www.ncbi.nlm.nih.gov/pmc/articles/PMC10145690/},
	doi = {10.3390/jcdd10040175},
	
	number = {4},
	urldate = {2025-03-06},
	journal = {J Cardiovasc Dev Dis},
	author = {Martínez-Sellés, Manuel and Marina-Breysse, Manuel},
	month = apr,
	year = {2023},
	pmid = {37103054},
	pmcid = {PMC10145690},
	pages = {175},
}

@article{mcdermott_reproducibility_2021,
	title = {Reproducibility in machine learning for health research: {Still} a ways to go},
	volume = {13},
	shorttitle = {Reproducibility in machine learning for health research},
	url = {https://www.science.org/doi/10.1126/scitranslmed.abb1655},
	doi = {10.1126/scitranslmed.abb1655},
	number = {586},
	urldate = {2025-03-06},
	journal = {Science Translational Medicine},
	author = {McDermott, Matthew B. A. and Wang, Shirly and Marinsek, Nikki and Ranganath, Rajesh and Foschini, Luca and Ghassemi, Marzyeh},
	month = mar,
	year = {2021},
	note = {Publisher: American Association for the Advancement of Science},
	pages = {eabb1655},
}

@article{jang_effectiveness_2021,
	title = {Effectiveness of {Transfer} {Learning} for {Deep} {Learning}-{Based} {Electrocardiogram} {Analysis}},
	volume = {27},
	url = {https://synapse.koreamed.org/articles/1146459},
	doi = {10.4258/hir.2021.27.1.19},
	number = {1},
	urldate = {2025-03-06},
	journal = {Healthc Inform Res},
	author = {Jang, Jong-Hwan and Kim, Tae Young and Yoon, Dukyong},
	month = jan,
	year = {2021},
	note = {Publisher: Korean Society of Medical Informatics},
	pages = {19--28},
}

@article{chato_survey_2023,
	title = {Survey of {Transfer} {Learning} {Approaches} in the {Machine} {Learning} of {Digital} {Health} {Sensing} {Data}},
	volume = {13},
	copyright = {http://creativecommons.org/licenses/by/3.0/},
	issn = {2075-4426},
	url = {https://www.mdpi.com/2075-4426/13/12/1703},
	doi = {10.3390/jpm13121703},
	
	language = {en},
	number = {12},
	urldate = {2025-03-06},
	journal = {Journal of Personalized Medicine},
	author = {Chato, Lina and Regentova, Emma},
	month = dec,
	year = {2023},
	note = {Number: 12
Publisher: Multidisciplinary Digital Publishing Institute},
	pages = {1703},
}

\section*{Conflict of Interest}
The authors declare no competing interests.
\section*{Author Contributions - CRediT}
\textbf{Lucas Bickmann}: Conceptualization, Methodology, Software, Data Curation, Writing - Original Draft, Visualization
\textbf{Lucas Plagwitz}: Formal analysis, Writing - Review \& Editing, Visualization
\textbf{Antonius Büscher}: Resources, Data Curation, Writing - Review \& Editing
\textbf{Lars Eckardt}: Conceptualization, Formal analysis, Writing - Review \& Editing, Supervision
\textbf{Julian Varghese}: Conceptualization, Resources, Writing - Review \& Editing, Supervision, Project administration, Funding acquisition
\section*{Data \& Code Availability}
All Electrocardiogram datasets are available under Physionet or their respective publication, except EDMS which is not yet openly published. The code of ExChanGeAI and CardX are open-source, and based on the MIT License. The end-to-end platform ExChanGeAI and the foundation model CardX are available under \url{imigitlab.uni-muenster.de/published/exchangeai}.

\newpage
\appendix
\onecolumn
\vspace*{1pt}
\section{Supplement}
\startsupplement
\begin{figure}[H]
    \centering
    \includegraphics[width=0.9\linewidth]{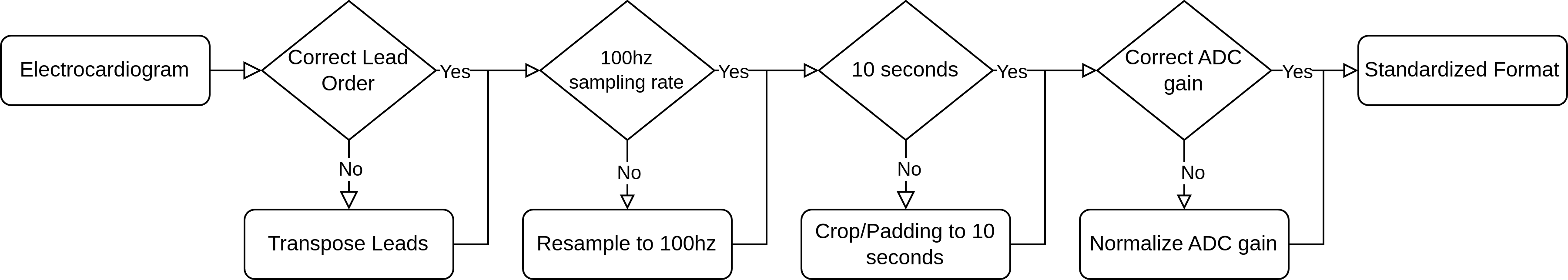}
    \caption{Flowchart of the pre-processing applied to any ECG data while being loaded into the application, independent of the file format.}
    \label{fig:norm-flowchart}
\end{figure}
\begin{table}[H]
\centering
\begin{tabular}{|l|cccc|cc|ccc|}
\hline
\textit{} &
  \multicolumn{4}{c|}{Superclasses} &
  \multicolumn{2}{c|}{Myocardial infarcts} &
  \multicolumn{3}{c|}{Bundle branch blocks} \\ \hline
\multicolumn{1}{|c|}{Stratified Fold} &
  \multicolumn{1}{c|}{MI} &
  \multicolumn{1}{c|}{CD} &
  \multicolumn{1}{c|}{STTC} &
  HYP &
  \multicolumn{1}{c|}{AMI} &
  IMI &
  \multicolumn{1}{c|}{CLBBB} &
  \multicolumn{1}{c|}{CRBBB} &
  IRBBB \\ \hline
1-8 (pretraining) &
  \multicolumn{1}{c|}{2.043} &
  \multicolumn{1}{c|}{1.353} &
  \multicolumn{1}{c|}{1.905} &
  415 &
  \multicolumn{1}{c|}{576} &
  984 &
  \multicolumn{1}{c|}{285} &
  \multicolumn{1}{c|}{75} &
  257 \\ \hline
Total &
  \multicolumn{4}{c|}{5.716} &
  \multicolumn{2}{c|}{1.560} &
  \multicolumn{3}{c|}{617} \\ \hline
9 (finetuning) &
  \multicolumn{1}{c|}{233} &
  \multicolumn{1}{c|}{171} &
  \multicolumn{1}{c|}{254} &
  64 &
  \multicolumn{1}{c|}{58} &
  123 &
  \multicolumn{1}{c|}{30} &
  \multicolumn{1}{c|}{19} &
  18 \\ \hline
Total &
  \multicolumn{4}{c|}{722} &
  \multicolumn{2}{c|}{181} &
  \multicolumn{3}{c|}{67} \\ \hline
10 (test) &
  \multicolumn{1}{c|}{256} &
  \multicolumn{1}{c|}{184} &
  \multicolumn{1}{c|}{242} &
  56 &
  \multicolumn{1}{c|}{66} &
  139 &
  \multicolumn{1}{c|}{38} &
  \multicolumn{1}{c|}{12} &
  36 \\ \hline
Total &
  \multicolumn{4}{c|}{738} &
  \multicolumn{2}{c|}{205} &
  \multicolumn{3}{c|}{86} \\ \hline
\end{tabular}
\caption{Dataset and class distribution for training samples across defined categories. Fold 1-8 is used for pretraining, fold 9 for fine-tuning and fold 10 for testing.}
\label{supp:datasets}
\end{table}
\begin{table}[H]
\centering
\begin{tabular}{|ll|cccccccc|}
\hline
\multicolumn{2}{|l|}{Model} &
  \multicolumn{2}{c|}{XceptionTime} &
  \multicolumn{1}{c|}{InceptionTime} &
  \multicolumn{1}{c|}{CardX‡} &
  \multicolumn{2}{c|}{DSAIL SNU} &
  \multicolumn{2}{c|}{ECG-FM} \\ \hline
\multicolumn{2}{|l|}{Task} &
  \multicolumn{3}{c|}{Training} &
  \multicolumn{2}{c|}{Finetune} &
  \multicolumn{2}{c|}{\textit{Physionet 2021}} &
  Finetune \\ \hline
\multicolumn{2}{|l|}{Finetuned PTB-XL folds} &
  \multicolumn{1}{c|}{1-8} &
  \multicolumn{4}{c|}{9} &
  \multicolumn{2}{c|}{-} &
  9 \\ \hline
\multicolumn{2}{|l|}{Finetuned layers} &
  \multicolumn{3}{c|}{all} &
  \multicolumn{2}{c|}{head} &
  \multicolumn{2}{c|}{-} &
  head \\ \hline
\multicolumn{2}{|l|}{Superclasses} &
  \multicolumn{1}{c|}{\textit{.792}} &
  \multicolumn{1}{c|}{{\ul .686}} &
  \multicolumn{1}{c|}{.651} &
  \multicolumn{1}{c|}{.520} &
  \multicolumn{1}{c|}{.536} &
  \multicolumn{1}{c|}{\textit{.038}} &
  \multicolumn{1}{c|}{\textit{.174}} &
  \textbf{.690} \\ \hline
\multicolumn{2}{|l|}{Myocardial infarcts} &
  \multicolumn{1}{c|}{\textit{.938}} &
  \multicolumn{1}{c|}{{\ul .853}} &
  \multicolumn{1}{c|}{\textbf{.902}} &
  \multicolumn{1}{c|}{.730} &
  \multicolumn{1}{c|}{.685} &
  \multicolumn{2}{c|}{\textit{-}} &
  .685 \\ \hline
\multicolumn{2}{|l|}{Bundle branch blocks} &
  \multicolumn{1}{c|}{\textit{.911}} &
  \multicolumn{1}{c|}{\textbf{.912}} &
  \multicolumn{1}{c|}{{\ul .891}} &
  \multicolumn{1}{c|}{.746} &
  \multicolumn{1}{c|}{.832} &
  \multicolumn{1}{c|}{\textit{.000}} &
  \multicolumn{1}{c|}{\textit{.016}} &
  .790 \\ \hline
\end{tabular}
\caption{Weighted F1 score for finetuned models on the internal PTB-XL test-dataset.}
\label{supp:weightedf1-internal}
\end{table}
\begin{table}[H]
\centering
\begin{tabular}{|l|cccccc|}
\hline
Model &
  \multicolumn{2}{c|}{XceptionTime} &
  \multicolumn{1}{c|}{InceptionTime} &
  \multicolumn{1}{c|}{CardX} &
  \multicolumn{1}{c|}{DSAIL SNU} &
  ECG-FM \\ \hline
Task                   & \multicolumn{3}{c|}{Training}  & \multicolumn{3}{c|}{Finetune} \\ \hline
Finetuned PTB-XL folds & \multicolumn{1}{c|}{1-8} & \multicolumn{5}{c|}{9}              \\ \hline
Finetuned layers       & \multicolumn{3}{c|}{all}       & \multicolumn{3}{c|}{head}     \\ \hline
Superclasses &
  \multicolumn{1}{c|}{\textit{.768}} &
  \multicolumn{1}{c|}{\textbf{.662}} &
  \multicolumn{1}{c|}{{\ul .617}} &
  \multicolumn{1}{c|}{.401} &
  \multicolumn{1}{c|}{.502} &
  .555 \\ \hline
Myocardial infarcts &
  \multicolumn{1}{c|}{\textit{.930}} &
  \multicolumn{1}{c|}{{\ul .827}} &
  \multicolumn{1}{c|}{\textbf{.888}} &
  \multicolumn{1}{c|}{.691} &
  \multicolumn{1}{c|}{.649} &
  .645 \\ \hline
Bundle branch blocks &
  \multicolumn{1}{c|}{\textit{.876}} &
  \multicolumn{1}{c|}{\textbf{.875}} &
  \multicolumn{1}{c|}{{\ul .853}} &
  \multicolumn{1}{c|}{.656} &
  \multicolumn{1}{c|}{.765} &
  .721 \\ \hline
\end{tabular}
\caption{Macro F1 score for finetuned models on the internal PTB-XL test datasets, excluding Physionet 2021 models.}
\label{supp:macrof1-internal}
\end{table}
\begin{table}[H]
\vspace*{1em}
\centering
\begin{tabular}{|ll|ccc|ccc|}
\hline
\multicolumn{2}{|l|}{Model} &
  \multicolumn{2}{c|}{XceptionTime} &
  InceptionTime &
  \multicolumn{1}{c|}{CardX} &
  \multicolumn{1}{c|}{DSAIL SNU} &
  ECG-FM \\ \hline
\multicolumn{1}{|l|}{Targets} &
  \multicolumn{1}{c|}{Dataset} &
  \multicolumn{1}{c|}{baseline} &
  \multicolumn{2}{c|}{all} &
  \multicolumn{3}{c|}{head} \\ \hline
\multicolumn{1}{|l|}{\multirow{3}{*}{Superclasses}} &
  Yang et al. &
  \multicolumn{1}{c|}{.112} &
  \multicolumn{1}{c|}{\textbf{.430}} &
  .021 &
  \multicolumn{1}{c|}{.171} &
  \multicolumn{1}{c|}{{\ul .242}} &
  .070 \\ \cline{2-8} 
\multicolumn{1}{|l|}{} &
  MIMIC-IV &
  \multicolumn{1}{c|}{.278} &
  \multicolumn{1}{c|}{\textbf{.280}} &
  {\ul .269} &
  \multicolumn{1}{c|}{.117} &
  \multicolumn{1}{c|}{.243} &
  .136 \\ \cline{2-8} 
\multicolumn{1}{|l|}{} &
  EDMS &
  \multicolumn{1}{c|}{.293} &
  \multicolumn{1}{c|}{{\ul .266}} &
  \textbf{.379} &
  \multicolumn{1}{c|}{.143} &
  \multicolumn{1}{c|}{.219} &
  .108 \\ \hline
\multicolumn{1}{|l|}{\multirow{2}{*}{Myocardial infarcts}} &
  MIMIC-IV &
  \multicolumn{1}{c|}{.753} &
  \multicolumn{1}{c|}{\textbf{.734}} &
  {\ul .726} &
  \multicolumn{1}{c|}{.499} &
  \multicolumn{1}{c|}{.336} &
  .410 \\ \cline{2-8} 
\multicolumn{1}{|l|}{} &
  EDMS &
  \multicolumn{1}{c|}{.546} &
  \multicolumn{1}{c|}{.469} &
  {\ul .570} &
  \multicolumn{1}{c|}{\textbf{.630}} &
  \multicolumn{1}{c|}{.352} &
  .495 \\ \hline
\multicolumn{1}{|l|}{\multirow{3}{*}{Bundle branch blocks}} &
  Yang et al. &
  \multicolumn{1}{c|}{.548} &
  \multicolumn{1}{c|}{.282} &
  \textbf{.648} &
  \multicolumn{1}{c|}{.276} &
  \multicolumn{1}{c|}{{\ul .404}} &
  .306 \\ \cline{2-8} 
\multicolumn{1}{|l|}{} &
  MIMIC-IV &
  \multicolumn{1}{c|}{.825} &
  \multicolumn{1}{c|}{\textbf{.820}} &
  \textbf{.819} &
  \multicolumn{1}{c|}{{\ul .439}} &
  \multicolumn{1}{c|}{.333} &
  .166 \\ \cline{2-8} 
\multicolumn{1}{|l|}{} &
  EDMS &
  \multicolumn{1}{c|}{.741} &
  \multicolumn{1}{c|}{\textbf{.821}} &
  {\ul .733} &
  \multicolumn{1}{c|}{\textbf{.820}} &
  \multicolumn{1}{c|}{.630} &
  .302 \\ \hline
\multicolumn{1}{|l|}{Revascularization} &
  EDMS &
  \multicolumn{1}{c|}{.750} &
  \multicolumn{1}{c|}{\textbf{.688}} &
  .546 &
  \multicolumn{1}{c|}{.614} &
  \multicolumn{1}{c|}{{\ul .635}} &
  .603 \\ \hline
\multicolumn{1}{|l|}{\multirow{4}{*}{Macro F1}} &
  Average↑ &
  \multicolumn{1}{c|}{.538} &
  \multicolumn{1}{c|}{\textbf{.532}} &
  {\ul .523} &
  \multicolumn{1}{c|}{.412} &
  \multicolumn{1}{c|}{.377} &
  .288 \\ \cline{2-8} 
\multicolumn{1}{|l|}{} &
  Median↑ &
  \multicolumn{1}{c|}{.645} &
  \multicolumn{1}{c|}{{\ul .579}} &
  {\ul \textbf{.609}} &
  \multicolumn{1}{c|}{.469} &
  \multicolumn{1}{c|}{.344} &
  .304 \\ \cline{2-8} 
\multicolumn{1}{|l|}{} &
  IQR↓ &
  \multicolumn{1}{c|}{.457} &
  \multicolumn{1}{c|}{.452} &
  .347 &
  \multicolumn{1}{c|}{.443} &
  \multicolumn{1}{c|}{\textbf{.161}} &
  {\ul .274} \\ \cline{2-8} 
\multicolumn{1}{|l|}{} &
  CV↓ &
  \multicolumn{1}{c|}{.475} &
  \multicolumn{1}{c|}{{\ul .442}} &
  .491 &
  \multicolumn{1}{c|}{.606} &
  \multicolumn{1}{c|}{\textbf{.416}} &
  .643 \\ \hline
\end{tabular}
\caption{Macro F1 score for finetuned models on the test datasets, excluding Physionet 2021 models.}
\label{supp:macrof1-external}
\end{table}
\begin{table}[H]
\centering
\begin{tabular}{|l|cc|ccc|}
\hline
               & \multicolumn{2}{c|}{Superclasses} & \multicolumn{3}{c|}{Bundle branch blocks}                       \\ \hline
PTB-XL classes & \multicolumn{1}{c|}{CD}   & STTC  & \multicolumn{1}{c|}{CLBBB} & \multicolumn{1}{c|}{CRBBB} & IRBBB \\ \hline
Physionet classes &
  \multicolumn{1}{p{4.5cm}|}{BBB, IRBBB, IAVB, LAnFB, CLBBB$|$LBBB,  CRBBB$|$RBBB, NSIVCB} &
  Ab, TInv, LQT &
  \multicolumn{1}{c|}{CLBBB$|$LBBB} &
  \multicolumn{1}{c|}{CRBBB$|$RBBB} &
  IRBBB \\ \hline
EDMS classes   & \multicolumn{2}{c|}{-}            & \multicolumn{1}{c|}{LBBB}  & \multicolumn{2}{c|}{RBBB}          \\ \hline
\end{tabular}
\caption{Physionet and EDMS to PTB-XL Classes matching for superclasses and bundle branch blocks.}
\label{supp:class-matching}
\end{table}
\subsection*{Carbon Footprint}
Experiments were conducted with private infrastructure, which has an estimated carbon efficiency of approximately 0.275kg $CO_2$ eq/kWh. A cumulative of 72 hours of computation was performed on hardware of type Nvidia A100 40GB (TDP of 250W). Total emissions are estimated to be 4.63 kg $CO_2$ equivalent for a single training of the foundation model CardX.
\end{document}